%% file: main.tex
\documentclass[10pt,twocolumn,letterpaper]{article}

\usepackage[pagenumbers]{cvpr} 
\makeatletter
\@namedef{ver@everyshi.sty}{}
\usepackage{graphicx}
\usepackage{amsmath}
\usepackage{amssymb}
\usepackage{booktabs}
\usepackage[accsupp]{axessibility} 
\usepackage{changes}
\usepackage[section]{placeins}
\usepackage[pagebackref,breaklinks,colorlinks]{hyperref}

\usepackage[capitalize]{cleveref}
\crefname{section}{Sec.}{Secs.}
\Crefname{section}{Section}{Sections}
\Crefname{table}{Table}{Tables}
\crefname{table}{Tab.}{Tabs.}

\def\cvprPaperID{12114} 
\def\confName{CVPR}
\def\confYear{2023}

\input{math_commands.tex}

\usepackage{caption}
\usepackage{xcolor}
\usepackage{hyperref}
\usepackage{url}
\usepackage{graphicx}
\usepackage{multirow}
\usepackage{diagbox}
\usepackage{pifont}
\usepackage{wrapfig}
\usepackage{soul}
\newcommand{\cmark}{\ding{51}}%
\newcommand{\xmark}{\ding{55}}
\newcommand{\sysname}{Img2LLM}

\definecolor{ForestGreen}{RGB}{5,166,88}
\definecolor{LavaRed}{RGB}{222,48,28}
\definecolor{LightGrey}{RGB}{180,180,180}

\begin{document}

\title{ From Images to Textual Prompts: Zero-shot Visual Question Answering with Frozen Large Language Models}

\author{Jiaxian Guo$^1$\thanks{Work done while Jiaxian Guo was an intern at Salesforce Research.}  \qquad Junnan Li$^2$  \qquad Dongxu Li$^2$ \qquad Anthony Meng Huat Tiong$^{2,3}$ \\  Boyang Li$^{3}$ \qquad Dacheng Tao$^{1}$ \qquad Steven Hoi$^{2}$\\
 $^1$ The University of Sydney  \qquad $^2$ Salesforce Research \qquad $^3$ Nanyang Technological University  \\
\tt \small jguo5934@uni.sydney.edu.au \quad \{junnan.li,li.d,anthony.tiong,shoi\}@salesforce.com \\
\tt \small boyang.li@ntu.edu.sg \quad dacheng.tao@gmail.com}
\maketitle

\begin{abstract}

\input{0-abstract.tex}
\end{abstract}

\section{Introduction}

\input{1-intro.tex}
\section{Related Work} \label{sec:related}
\input{2-preliminaries.tex}

\section{Method}

\input{3-method.tex}

\section{Experiment}
\input{4-exp.tex}

\section{Limitation}
\input{5-limitation.tex}
\section{Conclusion}
\input{6-conc.tex}
\section{Acknowledgments}
Jiaxian Guo was supported  in part by Australian Research Council Projects FL170100117 and IH180100002, the University of Sydney Completion Stipend Scholarship, and Faculty of Engineering PhD Completion Award. Boyang Li was supported by the Nanyang Associate Professorship and the National Research Foundation Fellowship (NRF-NRFF13-2021-0006), Singapore. 

{\small
\bibliographystyle{ieee_fullname}
\bibliography{iclr2023_conference}
}

\input{7-appendix.tex}

\end{document}

%% file: math_commands.tex
\usepackage{amsmath,amsfonts,bm}

\def\eqref#1{equation~\ref{#1}}

\def\1{\bm{1}}

\DeclareMathAlphabet{\mathsfit}{\encodingdefault}{\sfdefault}{m}{sl}
\SetMathAlphabet{\mathsfit}{bold}{\encodingdefault}{\sfdefault}{bx}{n}

\newcommand{\R}{\mathbb{R}}

\DeclareMathOperator*{\argmax}{arg\,max}

%% file: 0-abstract.tex
Large language models (LLMs) have demonstrated excellent zero-shot generalization to new language tasks. However, effective utilization of LLMs for zero-shot visual question-answering (VQA) remains challenging, primarily due to the modality disconnect and task disconnect between the LLM and VQA tasks. 
End-to-end training on multimodal data may bridge the disconnects, but is inflexible and computationally expensive. To address this issue, we propose \emph{Img2LLM}, a plug-and-play module that provides LLM prompts to enable LLMs to perform zero-shot VQA tasks without end-to-end training. We develop LLM-agnostic models describe image content as exemplar question-answer pairs, which prove to be effective LLM prompts.
Img2LLM offers the following benefits: 1) It achieves comparable or better performance than methods relying on end-to-end training. For example, we outperform Flamingo~\cite{Deepmind:Flamingo2022} by 5.6\% on VQAv2.
On the challenging A-OKVQA dataset, our method outperforms few-shot methods by as much as 20\%.
2) It flexibly interfaces with a wide range of LLMs to perform VQA. 3) It eliminates the need to specialize LLMs using end-to-end finetuning and serve highly specialized LLMs to end users, thereby reducing cost.
%
Code is available via the LAVIS~\cite{lavis} framework at \url{https://github.com/salesforce/LAVIS/tree/main/projects/img2llm-vqa}.





%% file: 1-intro.tex
Visual question answering (VQA) \cite{antol2015vqa} is a prominent vision-language task that finds a broad range of real-world applications, such as assisting blind individuals in understanding their environments. A diverse set of VQA datasets have been proposed, some focusing on image recognition \cite{goyal2017making,antol2015vqa} and others on logical reasoning \cite{marino2019ok}. 
However, human annotations are expensive to obtain and may introduce a variety of human biases \cite{changpinyo2022all,banerjee2020weaqa,yuan2021language}, making the VQA system brittle towards new answer styles and question types\cite{agrawal2018don,kafle2017analysis}. This has led researchers to zero-shot VQA methods \cite{changpinyo2022all,banerjee2020weaqa,kafle2017analysis} that do not require ground-truth question-answer annotations, thereby facilitating more generalizable VQA systems.


Recently, large language models (LLMs) (e.g., \cite{GPT-3,zhang2022opt}) have demonstrated excellent capabilities to perform tasks with zero in-domain data, conduct logical reasoning, and apply commonsense knowledge in NLP tasks \cite{kojima2022large,wei2022emergent,wei2022chain}. As a result, recent approaches \cite{Deepmind:Flamingo2022,pica:yang2021empirical,frozen:tsimpoukelli2021multimodal} have resorted to leverage LLMs in zero-shot VQA. 

However, applying LLMs to VQA tasks is less than straightforward, due to (1) the modality disconnect between vision and language and (2) the task disconnect between language modeling and question answering. A common technique is to finetune a vision encoder jointly with the LLM \cite{frozen:tsimpoukelli2021multimodal,Deepmind:Flamingo2022,fewvlm:jin2021good} to align the vision and language representation spaces, but this can incur prohibitive computational and data cost. For example, Flamingo \cite{Deepmind:Flamingo2022} finetunes on billions of image-text pairs with thousands of TPUs. Further, the finetuning specializes and introduces strong interdependence between the vision encoder and the LLM. If we need to upgrade the LLM as new versions emerge, the entire model needs to undergo expensive re-training.

In contrast to the end-to-end integration of LLM into a VQA system, this paper proposes a modular VQA system built on top of frozen off-the-shelf LLMs. This brings two benefits. First, it can reduce the deployment cost and simplify the deployment. 
Second, upgrading the LLM is straightforward. However,
it is challenging to bridge the modality disconnect and task disconnect without end-to-end training.
PICa~\cite{pica:yang2021empirical} converts images into captions, and provides exemplar QA pairs from training data as prompt to the LLM. However, doing so assumes the existence of annotated training data and the performance is sensitive to the selection of few-shot exemplars.

We propose \emph{\sysname}, a plug-and-play module that enables off-the-shelf LLMs to perform zero-shot VQA.  The central insight of \sysname{} is that we can utilize a vision-language model (\emph{e.g.} BLIP \cite{li2022blip}) and a question-generation model to translate the image content into synthetic question-answer (QA) pairs, which are fed to the LLM as part of the prompt. These exemplar QA pairs tackle the modality disconnect by describing the image content verbally, and tackle the task disconnect by demonstrating the QA task to the LLM. Notably, the exemplar QA pairs are constructed entirely based on the test image and question, obviating the need for similar few-shot examples as required by PICa \cite{pica:yang2021empirical}, which are not always available in practical zero-shot scenarios.
When applied to the open-source OPT language models \cite{zhang2022opt}, Img2LLM achieves comparable or superior zero-shot VQA performance to methods that perform costly end-to-end training.

With this paper, we make the following contributions.

\begin{itemize}
 \vspace{-0.2em}
    \item 
    We propose Img2LLM, a plug-and-play module that converts an image into synthetic question-answer pairs based solely on the current image of the question.
    Img2LLM bridges the modality disconnect between language and vision as well as the task disconnect between language modeling and visual question-answering.
     \vspace{-0.2em}
    \item Img2LLM enables off-the-shelf LLMs to perform zero-shot VQA without costly end-to-end training or specialized textual QA networks \cite{meng2022plug}, thereby allowing low-cost and flexible model deployment and painless LLM upgrades (Table \ref{tab:llms}).
 \vspace{-0.2em}
\item Our experimental results show that the OPT models equipped with Img2LLM achieve zero-shot VQA performance that is competitive or superior to the end-to-end trained models. 
For example, we outperform Flamingo~\cite{Deepmind:Flamingo2022} by 5.6\% on VQAv2.
We even outperform many few-shot VQA methods.
 \vspace{-0.2em}
\end{itemize}

%% file: 2-preliminaries.tex




\subsection{Recent Advances in VQA Methods}
As a multi-modal evaluation benchmark, Visual Question Answering (VQA) that
requires the model to answer a natural language question according to the image, has been the focus of active
research \cite{yang2016stacked,anderson2018bottom,antol2015vqa,schwenk2022okvqa,akula2021crossvqa}.
The past few years witnessed rapid performance advances with large-scale image-text pretraining \cite{jiang2022pseudo,yuan2021florence,lu2019vilbert,li2021align,li2020oscar,zhang2021vinvl,wang2021simvlm,singh2021flava,li2022blip,fewvlm:jin2021good,dai2022enabling} followed byfine-tuning on VQA datasets.
To tackle knowledge-based VQA \cite{schwenk2022okvqa,marino2019ok}, recent works \cite{gui2021kat,lin2022revive,wu2022multi,luo2022vc,luo2021weakly,marino2021krisp,garderes2020conceptbert,li2020boosting} incorporate external knowledge, such as ConceptNet \cite{speer2017conceptnet} or Wikipedia, but experimental results in \cite{schwenk2022okvqa} show that these methods still struggle to answer questions requiring complex reasoning.

\subsection{LLM for Zero/Few-Shot VQA Tasks}
\label{sec:llm-vqa}
Large language models (LLMs)~\cite{brown2020language,zhang2022opt,chowdhery2022palm} trained on web-scale corpus are powerful in natural language understanding and reasoning~\cite{Zhou-2022:Least-to-Most-Prompting,GPT-3}.
To infer on task data, LLMs typically generate target tokens autoregressively.
%
%
In specific, given prompt $C$ and task input $x$, an LLM generates target tokens $Y =\{y_i\}_{i=1}^n$, with
$y_i$ = $\displaystyle \argmax p_{\theta}(y_i|y_{<i},C,x) $ and $\theta$ the model parameters.
Prior VQA methods using LLMs mainly fall into two categories: multi-modal pretraining and language-mediated VQA.

\textbf{Multi-modal pretraining}.~~These approaches align vision and language embeddings by training additional alignment modules, as shown in Figure \ref{fig:pardiam_illustration}(a). 
Considering that LLMs are too large to finetune efficiently, \cite{frozen:tsimpoukelli2021multimodal} opt to fine-tune only the visual encoder while Flamingo \cite{Deepmind:Flamingo2022} trains extra cross-attention layers to model cross-modality interactions.
%
However, this paradigm suffers from two drawbacks: 1) Highly compute-inefficient. Jointly aligning vision backbones and LLMs requires large compute resources. For example, training Flamingo requires 1536 TPUv4 over two weeks. Hence, it becomes prohibitively expensive to switch to a different LLM.
2) Catastrophic forgetting. The alignment step may be detrimental to LLMs' reasoning ability, if the LLMs are jointly trained with the visual model~\cite{Deepmind:Flamingo2022}.
%


\textbf{Language-mediated VQA}.~~Instead of vectorized representations, this VQA paradigm directly resorts to natural language as the intermediate representation of the image and no longer requires expensive pretraining. 
As depicted by Figure \ref{fig:pardiam_illustration}(b), it first converts the current image to language descriptions and feeds the descriptions, possibly accompanied by in-context exemplars, to a frozen LLM. 
In a few-shot setting, PICa \cite{pica:yang2021empirical} generates captions for the image and selects training data samples as in-context exemplars, but its performance degrades substantially when the exemplars are omitted. As a concurrent zero-shot approach, \cite{meng2022plug} generates question-relevant captions. Due to the zero-shot requirement, it is unable to provide in-context exemplars and does not reap the benefits of in-context learning. As a result, it has to rely on a QA-specific LLM, UnifiedQAv2 \cite{khashabi2020unifiedqa}, to achieve high performance.






%% file: 3-method.tex
\begin{figure*}[!t]
\vspace{-0.8em}
\centering
	  
			         \begin{minipage}[c]{\linewidth}
				             \centering

\centering
\includegraphics[width=\linewidth]{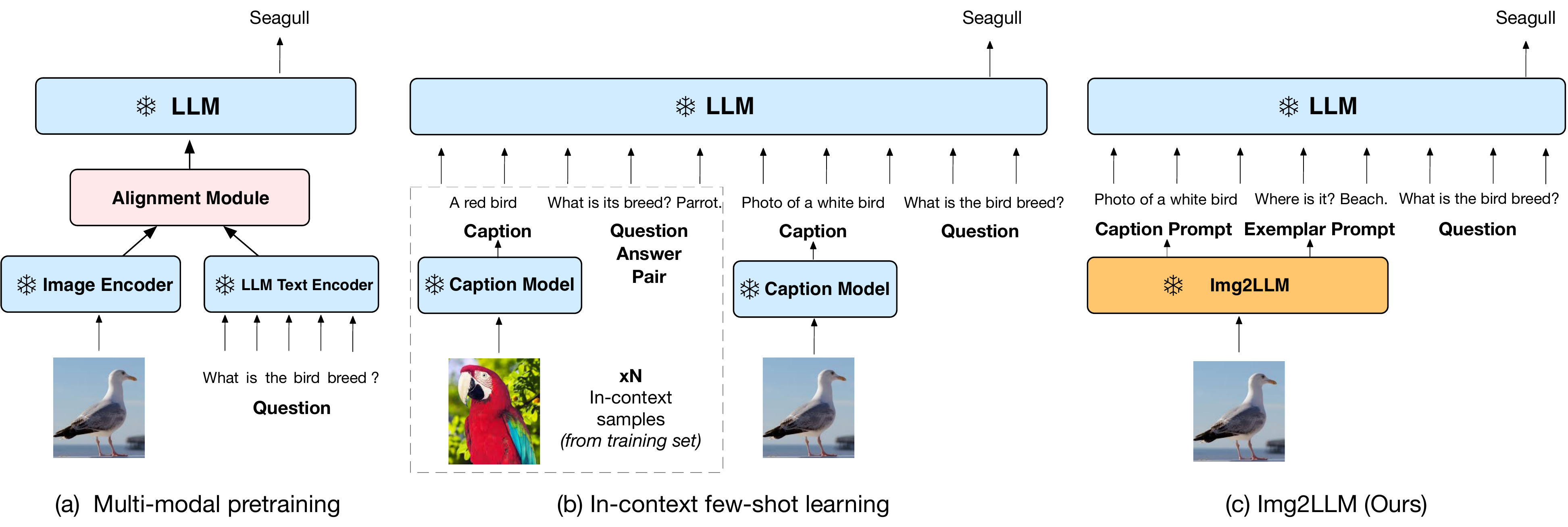}
				         \end{minipage}
			     
\vspace{-0.2em}
	\caption{{The illustrative comparison of three tyepes of methods that enable LLM to perform VQA tasks, where blue block denotes that the the inner parameters are frozen while pink block indicates the inner parameters are trainable. }}
 \vspace{-0.2em}
 \label{fig:pardiam_illustration}
 \label{fig:align}
\end{figure*}

Difficulties in utilizing LLMs effectively in zero-shot VQA stem mainly from two obstacles:
(i) \emph{The modality disconnection}: LLMs do not natively process images and encoding visual information into a format that LLMs can process can be a challenge.
(ii) \emph{The task disconnection}: LLMs are usually pretrained using generative~\cite{GPT-3} or denoising objectives~\cite{devlin2018bert} on language modeling tasks. As the LLMs are unaware of the tasks of question answering or VQA, they often fail to fully utilize contextual information in generating the answers. 





In language-mediated VQA \cite{pica:yang2021empirical,meng2022plug}, the modality disconnection is addressed by converting the image to intermediate language descriptions instead of dense vectors (\S \ref{sec:llm-vqa}). The task disconnection must be addressed using either few-shot in-context exemplars \cite{pica:yang2021empirical} or an LLM directly finetuned on textual QA \cite{meng2022plug}. It is not clear how to tackle the task disconnection on generic LLMs under zero-shot settings. 

We propose a new zero-shot technique to address the task disconnection on generic LLMs, \sysname{} (Figure~\ref{fig:pardiam_illustration}c), which generates image-relevant exemplar prompts for the LLM. Given a question $Q$ and an image, our key insight is that we can generate synthetic question-answer pairs as in-context exemplars from the \emph{current} image. The exemplars not only demonstrate the QA task but also communicate the content of the image to the LLM for answering the question $Q$, thereby hitting two birds with one stone. \sysname{} is LLM-agnostic; it unlocks the knowledge and the reasoning capacity of off-the-shelf LLMs, offering a powerful yet flexible solution for zero-shot VQA.




\subsection{Answer Extraction}\label{sec:ans_extract}
\begin{figure*}[!h]
\centering
	  
			         \begin{minipage}[c]{\linewidth}
				             \centering
				          
 \includegraphics[width=\textwidth]{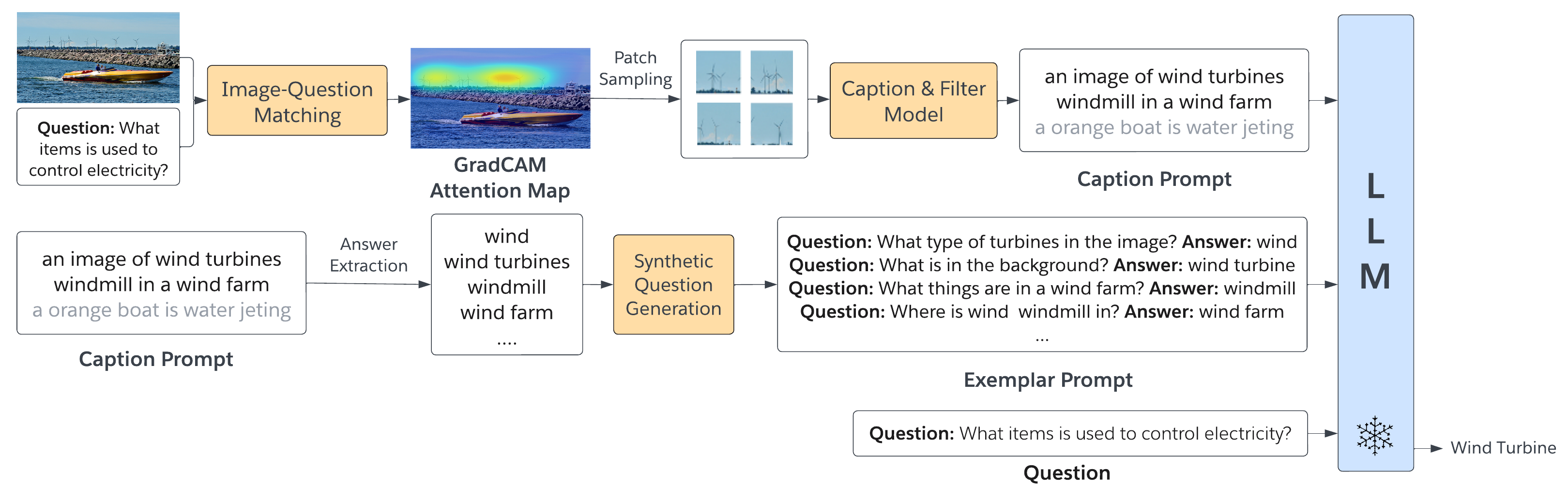}
				         \end{minipage}

 \vspace{-0.5em}
	\caption{{The overall pipeline of \sysname{}, including
Caption Prompt and Exemplar Prompt generation.} }
 \vspace{-0.3em}
 \label{fig:taskprompt}
\end{figure*}

 In order to incorporate the image content into the exemplars for in-context learning, from the current VQA image, we first seek words  that could serve as answers to synthetic questions. We generate a number of captions using an off-the-shelf question-relevant caption generation module (\S \ref{sec:qg_caption}). Following recent papers \cite{changpinyo2022all,lee2021qace}, we extract noun phrases (including named entities), verb phrases, adjective phrases, numbers, and boolean-typed words like ``yes'' and ``no'' as potential answers\footnote{We use the spaCy parser at \url{https://spacy.io/}, though are not tied to the parser in any way.}. We show some extracted answer candidates in Figure \ref{fig:taskprompt} and Appendix A.3.


\begin{table*}[!t]
\centering
\small
\setlength{\tabcolsep}{1.mm}
\caption{Results from mixing captions and exemplar prompts on 30B OPT \cite{zhang2022opt}.}
\label{tab:ablation}
\vspace{-0.3em}
\begin{tabular}{cccccccc}
\hline 
\setlength\tabcolsep{3pt}
Prompt Template& Caption Prompt  &     Exemplar Prompt&VQAv2 val &  OK-VQA\\
                                  \hline
Instruction&\xmark&\xmark & 18.1&3.3 \\
Instruction + Captions &\cmark&\xmark& 46.1& 23.5 \\
Instruction + Question-Answer Pairs &\xmark&\cmark& 57.9& 41.1 \\
Instruction + Captions + Question-Answer Pairs &\cmark&\cmark& 59.5& 41.8 \\
                                  \hline
\end{tabular}
\vspace{-0.3em}
\end{table*}

\subsection{Question Generation} \label{sec:q-select}
With the extracted answer candidate set $\{\hat{a}_j\}_{j=1}^U$, we can directly use any question generation network \cite{kafle2017data,lu202012,xu2020radial,kil2021discovering,akula2021crossvqa} to generate specific questions for each answer candidate. In this paper, we experiment with both template-based and neural question-generation methods. Note that to avoid violating the zero-shot requirements, our method is purely textual-based without access to any VQA data.

\textbf {Template-based Question Generation.} Using an off-the-shelf parser, we obtain the part-of-speech for each answer, and design specific question templates for each POS type. For example, for answers that are nouns, we use the question ``What object is in this image?'' For verb answers, we use the question ``What action is being taken in this image?'' Due to space constraints, we put the complete list of templates in Appendix A.5.


\textbf {Neural Question Generation.} Inspired by \cite{changpinyo2022all}, we train a neural question generation model on textual QA datasets. Specifically, we finetune a pretrained T5-large model \cite{Raffel:JMLR:T5} to generate questions from answers. The input to the model contains the prompt ``Answer: \texttt{[answer]}. Context: \texttt{[context]}'', where \texttt{[answer]} denotes the answer text and \texttt{[context]} denotes the context text from textual QA datasets. During inference, we replace \texttt{[answer]} with an extracted answer candidate and \texttt{[context]} with the generated caption from which the answer was extracted. 
The model is finetuned on five textual QA datasets including SQuAD2.0 \cite{rajpurkar2018know}, MultiRC \cite{khashabi2018looking}, BookQA \cite{mihaylov2018can}, CommonsenseQA \cite{talmor2018commonsenseqa} and Social IQA\cite{sap2019socialiqa}.

With the above question generation methods, we acquire a set of synthetic question-answer pairs $\{\hat{q}_j,\hat{a}_j\}_{j=1}^U$. We use these question-answer pairs as exemplars of LLM in-context learning \cite{GPT-3}, which guides the LLM to perform QA task given the image content and bridges the task disconnect between language modelling and VQA. 

As a sneak preview, we show effects of exemplar QA pairs in Table \ref{tab:ablation}. The details of the instructions are explained in \S \ref{sec:prompt_design}. We observe that exemplar QA prompts perform considerably better than caption prompts (detailed in \S \ref{sec:qg_caption}) only, demonstrating their efficacy in bridging the task disconnection between LLM pre-training and VQA tasks. Moreover, since the exemplar prompts already describe much content of the image, which helps to bridge the modality disconnection, adding captions on top does not provide much new information and brings only limited performance gains.

\subsection{Question-relevant Caption Prompt} \label{sec:qg_caption}
In addition to the synthetic exemplar QA pairs, we also supply question-relevant image captions to the LLM. 
We observe that the question may ask about specific objects or regions in the image \cite{wu2019generating} but generic captions generated by existing networks may not contain relevant information. In Figure \ref{fig:taskprompt}, the question \emph{``What items are spinning in the background which can be used to control electricity?"} is relevant only to the wind turbines. However, captions generated from the whole image are likely to focus on the salient orange boat, leaving LLM with no information to answer the question. To address this issue, we generate captions about the question-relevant portion of the image and include them in the prompt to the LLM.

To achieve this, we first determine the regions of the image that are relevant to the question, by using the Image-grounded Text Encoder (ITE) in BLIP \cite{li2022blip}, as which assigns a similarity score $\text{sim}(v, q)$ to any pair of image $v$ and textual question $q$. With ITE, we use GradCAM \cite{selvaraju2017grad}, a feature-attribution interpretability technique, to generate a coarse localisation map highlighting matching image regions given a question \cite{li2022blip}. Briefly, GradCam qualifies the cross-attention scores from the Transformer network by the gradient of ITE simlarity function $\text{sim}(v, q)$ with respect to the cross-attention scores. As this technique was proposed in \cite{meng2022plug}, we leave the details to Appendix A.1.

Having obtained the patch relevance $r$, we sample a subset of image patches with probability proportional to patch relevance $r$. After that, we generate captions from the sampled image patches using top-k sampling \cite{fan2018hierarchical}. To generate semantically meaningful captions, a short prompt, ``a picture of," is also fed into the text decoder. We repeat this $M$ times for each image to generate $M$ diverse captions, and keep only captions that are not exact substrings of others.
However, due to the non-deterministic nature of top-k sampling, the caption model may generate noisy captions that have a negative impact on performance. To remove noisy captions, we use ITE to calculate the similarity score between the generated caption and sampled question-relevant image patches, and filter captions with less than 0.5 matching scores. Overall, this process yields synthetic captions that are question-relevant, diverse, and clean, providing a bridge between visual and language information.

\subsection{Prompt Design} \label{sec:prompt_design}

With synthetic question-relevant captions and question-answer pairs, we construct complete prompts for LLM by concantenating 
the instruction, captions, and QA exemplars. The instruction text is ``Please reason the answers of question according to the contexts." The caption prompt is formatted as ``Contexts: \texttt{[all captions]}". Individual QA exemplars are formatted as ``Question: \texttt{[question]} Answer: \texttt{[answer]}'' and concatenated. We position the current question as the last portion of the prompt, formatted as ``Question: \texttt{[question]}. Answer:~''. Finally, to get the answer, we perform greedy decoding on the LLM and remove meaningless tokens as in Flamingo. 



Furthermore, as the input to LLMs has maximum lengths, \emph{e.g.} 2048 in OPT and GPT3, it is necessary to select a subset of question-relevant captions and question-answer pairs to construct the prompt. To select the most informative prompt, we first count the frequency of the synthetic answer candidates in 100 generated captions. We then select 30 answer candidates with highest frequencies and generate one question for each. Also, we include 30 answers with the lowest frequency and one caption containing each answer. 
See \S \ref{sec:cap-select} for analysis of caption selection strategies.



%% file: 4-exp.tex
In this section, we first validate the efficacy of \sysname{} by comparing it with other zero-shot and few-shot VQA methods.
%
Then, we perform ablation studies on important design choices, such as prompt patterns and caption selection strategies, to understand their effect.
We also show qualitative examples and include discussion on observed failure cases.



\subsection{Environment Setup} \label{sec:environemnt_setup}
\textbf{Datasets}. We validate our method on VQAv2 \cite{goyal2017making}, OK-VQA \cite{marino2019ok} and A-OKVQA \cite{schwenk2022okvqa} datasets, which contain questions requiring perception, reasoning and commonsense to answer. Specifically, VQAv2 \cite{goyal2017making} contains 214,354 questions in the validation set and 107,394 in the test-dev dataset. 
OK-VQA \cite{marino2019ok} and A-OK-VQA~\cite{schwenk2022okvqa} emphasize on commonsense reasoning, among which OK-VQA contains 5,046 test questions and A-OKVQA \cite{schwenk2022okvqa} contains 1,100 validation questions and 6,700 test questions.

\textbf{Implementation details}.  
To obtain question-relevant caption prompt, we use BLIP~\cite{li2022blip} to generate captions and perform image-question matching.
To localize the image regions relevant to the question, we generate GradCam from the cross-attention layer of BLIP image-grounded text encoder.
We then sample $K^\prime=20$ image patches based on GradCam, and use them to obtain 100 question-relevant captions.
For the LLMs, our main result uses the open-source OPT model with multiple different sizes.
Our ablation study also experiments with various other LLMs to show the generalization ability of our method.
%
%
We use LLMs to generate answers auto-regressively, without access to either answer list or training samples, thereby facilitating zero-shot VQA.
We follow official evaluation protocols and report VQA scores on each dataset.


%

\textbf{Competing methods}. We compare with prior VQA methods, which rougly fall into three categories: (i) \emph{Zero-shot methods with frozen LLMs}, such as PICa~\cite{pica:yang2021empirical}.
Our method also belongs to this category, yet unlike PICa, \sysname{} requires no training samples to compose the prompts.
(ii) \emph{Zero-shot methods with extra multi-modal pre-training}, such as Flamingo \cite{Deepmind:Flamingo2022}, Frozen \cite{frozen:tsimpoukelli2021multimodal}, VL-T5 \cite{cho2021unifying}, FewVLM \cite{fewvlm:jin2021good} and VLKD \cite{dai2022enabling}.
These methods require large-scale vision-language datasets and are costly to update.
We also include results from VQ\textsuperscript{2}A \cite{changpinyo2022all} and WeaQA \cite{banerjee2020weaqa} in this category, with \emph{caveats} that they assume access to answer candidates which may not be available in practice.
Therefore, their results should be interpreted with caution.
(iii) For reference purposes, we also include available results from \emph{few-shot methods}. 
These include few-shot results of PICa~\cite{pica:yang2021empirical}, FewVLM~\cite{fewvlm:jin2021good} and ClipCap~\cite{mokady2021clipcap}.
%


\subsection{Main Results} \label{sec:re}

\begin{table*}[!t]
\centering
\setlength{\tabcolsep}{1.8mm}\vspace{-0.5em}
\caption{Performance on VQAv2, OK-VQA, and A-OKVQA.
A few methods do not strictly satisfy the zero/few-shot requirements: methods without end-to-end training but assumes access to training samples are labeled with $\dagger$; methods that answer from a predefined list of candidates are in grey. Further, \textcolor{ForestGreen}{\xmark} annotates methods requiring no end-to-end training, which is desirable, and \textcolor{LavaRed}{\cmark} otherwise. 
}
\vspace{-0.5em}
\label{tab:main}
\begin{tabular}{ccccccccc}
\hline 
Methods &End-to-End  &Shot               & \multicolumn{2}{c}{VQAv2} & OK-VQA & \multicolumn{2}{c}{A-OKVQA}   \\
                              &Training? & Number& val & test & test & val    & test                                          \\
                                  \hline
                                   \multicolumn{8}{c} {\emph{ Zero-Shot Evaluation with Frozen Large Language Model}}  \\
PICa\textsubscript{175B}$^{\dagger}$                            &\textcolor{ForestGreen}{\xmark}&0& -     &- &17.7   &  - &-       \\
                            Img2LLM\textsubscript{6.7B}                                                    &\textcolor{ForestGreen}{\xmark}&0& 57.6  & 57.0 &38.2  & 33.3& 32.2 \\
                           Img2LLM\textsubscript{13B}                                                    &\textcolor{ForestGreen}{\xmark}&0& 57.1 & 57.3 &39.9  & 33.3&33.0 \\
                                  Img2LLM\textsubscript{30B}                                                    &\textcolor{ForestGreen}{\xmark}&0& 59.5  & 60.4 &41.8  & 36.9 &36.0 \\
Img2LLM\textsubscript{66B}                                                    &\textcolor{ForestGreen}{\xmark}&0& 59.9 &60.3 & 43.2  & 38.7 &38.2\\ 

                                   Img2LLM\textsubscript{175B}                                                    &\textcolor{ForestGreen}{\xmark}&0&  \textbf{60.6} &\textbf{61.9}  & 45.6  &\textbf{42.9} & \textbf{40.7}  \\
                                  \hline
                                   \multicolumn{8}{c} {\emph{Zero-Shot Evaluation with Extra End-to-End Training}} \\
                                   
                                    VL-T5\textsubscript{no-vqa}  &\textcolor{LavaRed}{\cmark} &0& 13.5& -    &5.8      &-  &-     \\
                                    FewVLM\textsubscript{base}   &\textcolor{LavaRed}{\cmark} & 0& 43.4&-        &11.6      &-&-          \\
                           FewVLM\textsubscript{large}  &\textcolor{LavaRed}{\cmark} &0& 47.7& -          &16.5&   -     &-       \\
                                        VLKD\textsubscript{ ViT-B/16} &\textcolor{LavaRed}{\cmark}&0& 38.6&39.7    &10.5        &-&-   \\
                                    VLKD\textsubscript{ ViT-L/14} &\textcolor{LavaRed}{\cmark}&0&42.6&44.5    &13.3      &-&-  \\
                            
                                    Frozen\textsubscript{7B} &\textcolor{LavaRed}{\cmark}&0& 29.5 &-& 5.9  &-\\
       Flamingo\textsubscript{3B}&\textcolor{LavaRed}{\cmark}  &0& -&49.2  & 41.2   &-   &-      \\
         Flamingo\textsubscript{9B} &\textcolor{LavaRed}{\cmark} & 0&-&51.8  & 44.7   &-  &-     \\
    Flamingo\textsubscript{80B} &\textcolor{LavaRed}{\cmark} & 0&-&56.3 &  \textbf{50.6}   &-&-         \\
                                 \hline
\multicolumn{8}{c} {\emph{Zero-shot Evaluation with Access to Answer Candidates}} \\
\textcolor{LightGrey}{WeaQA ZSL}&{\textcolor{LavaRed}{\cmark}}&\textcolor{LightGrey}{0}& \textcolor{LightGrey}{46.8}&-& -&-&-  \\
\textcolor{LightGrey}{VQ\textsuperscript{2}A} &{\textcolor{LavaRed}{\cmark}}&\textcolor{LightGrey}{0}&  \textcolor{LightGrey}{61.1} &- &\textcolor{LightGrey}{19.8} &-&-\\\hline
                                  \multicolumn{8}{c} {\emph{Few-Shot Evaluation}} \\
  ClipCap$\rightarrow$Cap$\rightarrow$GPT\textsubscript{175B} &\textcolor{ForestGreen}{\textcolor{ForestGreen}{\xmark}}&10& -&- &-&16.6 &15.8 \\
  ClipCap$\rightarrow$Rel$\rightarrow$GPT\textsubscript{175B} &\textcolor{ForestGreen}{\textcolor{ForestGreen}{\xmark}}&10&- &- &-&18.1&15.8 \\
FewVLM\textsubscript{base}   &\textcolor{LavaRed}{\cmark} & 16& 48.2&-     &    15.0  &-            \\
                           FewVLM\textsubscript{large}  &\textcolor{LavaRed}{\cmark} &16&  51.1&-        &  23.1    &-&-          \\
        PICa\textsubscript{175B}$^{\dagger}$    &\textcolor{ForestGreen}{\xmark}     & 1&     - &- &36.4 &-&-    \\PICa\textsubscript{175B}$^{\dagger}$       &\textcolor{ForestGreen}{\xmark}   &   4&    - &- &43.3 &-&-   \\
    PICa\textsubscript{175B}$^{\dagger}$   &\textcolor{ForestGreen}{\xmark}    &   16&   54.3 &- &46.5 &-&-   \\
    PICa\textsubscript{175B}-Ensemble   &\textcolor{ForestGreen}{\xmark}       &   80&    56.1 &- &48.0   &-&- \\
           \hline                  
\end{tabular}
\vspace{-1.2em}
\end{table*}

Main quantitative results are shown in Table \ref{tab:main}.
We summarize our findings as follows.

\textbf{State-of-the-art results on zero-shot evaluation with plug-in frozen LLMs.} \sysname{} surpasses PICa, the best prior zero-shot model with frozen LLMs, by a significant margin (45.6 \emph{versus} 17.7 on OK-VQA), thereby establishing a new state-of-the-art.
In addition, we remark that despite PICa uses frozen LLMs, it requires training samples to build prompts.
In contrast, our method generates question-answers with no access to VQA samples, thus fully fulfilling the zero-shot requirements.

\textbf{Scaling effect of LLMs and their emergent capabilities on VQA.}~When increasing the number of parameters of LLMs from 6.7B to 175B, we see a 3-10 points improvement in VQA across datasets.
This shows that stronger language modelling capabilities help better comprehend the question, thus giving more accurate answers.
Such a trend is more clear and consistent on OK-VQA and A-OKVQA, whose questions demand commonsense reasoning and external knowledge that LLMs excel at providing. This corroborates our belief that LLMs are beneficial to VQA.

Another intriguing phenomenon we observe is that the effect of scaling LLMs becomes obvious only when the model size becomes sufficiently large, for example, when using 30B or larger models, while not entirely predictable on smaller ones (6.7B and 13B).
This echoes with the recent finding on the emergent abilities when using LLMs off-the-shelf~\cite{emergent} for language tasks, while confirming the same trend for the first time when using frozen LLMs for vision(-language) tasks.

\textbf{Competitive performance with end-to-end pretraining and few-shot models.} \sysname{} obtains superior performance to most models with end-to-end pretraining,
as well as those evaluated in few-shot setups.
For example, on VQAv2 our method surpasses Flamingo$_{80\textnormal{B}}$, which cost over 500K TPU hours and billion-scale datasets to train, by a margin of 5.6 points.
On A-OKVQA, \sysname{} more than doubles the best reported results so far, from ClipClap.
The only a few exceptions are on OK-VQA, where our method obtains better results than Flamingo$_{9\textnormal{B}}$, yet is not able to stay on par with Flamingo$_{80\textnormal{B}}$.
Considering that \sysname{} is flexible to adapt to updated and stronger LLMs with zero extra training cost, we consider it a more approachable solution to practical adoption of VQA systems, than those trained end-to-end.
We also include comparisons with supervised models in Appendix A.4. 
\sysname{} achieves better performance than most supervised models, despite the fact that it uses zero training data and is evaluated in a zero-shot setup. These results once again validates its effectiveness.

\begin{table}[!htb]
\centering
\setlength{\tabcolsep}{1.mm}
\caption{Zero-shot VQA performance with different LLMs.}
\vspace{-0.5em}
\label{tab:llms}
\begin{tabular}{cccc}
\hline
Methods  & VQAv2 val & OK-VQA  \\
                          \hline
PICa\textsubscript{~GPT-3\ 175B}& - &
17.7\\ 
Frozen\textsubscript{7B}	&29.5	&5.9 \\
Ours\textsubscript{~GPT-Neo\ 2.7B}&  50.1 & 31.5\\ 
                            Ours\textsubscript{~BLOOM\ 7.1B}  & 52.4&
32.4    \\
                           Ours\textsubscript{~GPT-J\ 6B}&56.4&
37.4 \\
                            Ours\textsubscript{~OPT\ 6.7B}                                                   & 57.6&
38.2\\
Ours\textsubscript{~OPT\ 175B}                                                   & 60.6&
45.6\\
                                   \hline
\end{tabular}
\vspace{-1em}
\end{table}

\subsection{{Experimental Results of Different LLMs}}\label{app:llms}
In Table \ref{tab:llms}, we evaluate the performance of Img2LLM on various open-sourced LLMs other than OPT, including GPT-J \cite{gpt-j}, GPT-Neo \cite{gpt-neo} and BLOOM \cite{scao2022language}. 
The experimental results show that Img2LLM enables various LLMs to perform zero-shot VQA tasks, and that all of them achieve superior performance to zero-shot PICa~\cite{pica:yang2021empirical} and Frozen~\cite{frozen:tsimpoukelli2021multimodal}. This is a strong evidence for showing our method’s generalization ability with different LLMs.

\subsection{Analysis on Question Generation Methods} \label{sec:qg}

Table \ref{tab:question_selection} shows the performance of different question selection strategies described in Section \ref{sec:q-select}. We compare three question generation techniques, include image-\textit{agnostic}, which uses questions sampled from other images; \textit{template}-based, which uses template questions, and \textit{neural}-based, which uses neural generated questions. Further, we compare two synthetic QA selection strategies. The \textit{random} strategy, which selects QA pairs for prompt randomly; the \textit{max\ freq.} approach, which selects answer candidates that are most frequent in the captions, and also retrieve the associated synthetic questions to build the prompt. 

Among the three question generation techniques, \textit{Agnostic} perform the worst whereas \textit{Neural} performs the best. We attribute the differences to the quality of QA pairs. \textit{Agnostic} QA pairs contain information irrelevant to the current image and may mislead the LLM. \textit{Template} questions feature little linguistic variation and hence cannot demonstrate different QA strategies. \textit{Neural} has the most relevant information and the most linguistic diversity. 
QA pair with maximum answer frequency outperform random questions. We hypothesize that the most frequent answers describe the most salient or important aspects of the image, thereby providing more information than random questions.

\input{tables/table3-qselect.tex}

In addition, we evaluate visual information quality encoded in the exemplar prompts using the answer hit rate and the answer noise rate. Answer hit rate (AHR) is defined as the proportion of QA pairs containing the ground-truth answer. Answer noise rate (ANR) is defined as the ratio of ground-truth answers to the total number tokens in the exemplar prompts. Table \ref{tab:answer} indicates that exemplar prompts generated from question-relevant captions have a higher AHR, hence enhancing the VQA performance. In addition, the caption filter procedure can remove some noisy captions, allowing it to achieve a higher ANR than its competitors.  The experimental results demonstrate that improving both the AHR and the ANR can improve the quality of prompts and VQA performance.


\begin{table}[t]
\centering
 \makeatletter\def\@captype{table}\makeatother\caption{Ablations on prompts designs.}
  \vspace{-0.2em}
 \label{tab:cqa}
 \setlength{\tabcolsep}{1.5mm}
   \begin{tabular}{cccccc} 
\hline       Methods  &       OK-VQA&  VQAv2 val\\
                              \hline
CQA-CQA-CQA & 37.8 & 52.1\\
CCC-QAQAQA & 41.8 & 59.5  \\

                              \hline
\end{tabular}
\end{table}

  
  \begin{table}[t]
   \centering
        \makeatletter\def\@captype{table}\makeatother\caption{Ablation on caption selection methods.}
        \label{tab:caption_selection}
         \vspace{-0.2em}
        \setlength{\tabcolsep}{1.5mm}
         \begin{tabular}{cccc}        
          \hline 
                                                                            Caption&        \multirow{2}{*}{Random} & Max  & Min \\
Selection& &Frequency &Frequency
                                                                            \\
                                  \hline
 OK-VQA Acc  & 41.3&41.1 & \bf41.8 \\

                                  \hline
      \end{tabular}
      \vspace{-0.5em}
   \end{table}

\input{tables/table4.tex}

\subsection{Ablation on Caption Selection} \label{sec:cap-select}

As Table \ref{tab:caption_selection} shows, we evaluate the performance different caption selection strategies, where Max Frequency selects captions containing 30 answers with highest frequencies and Min Frequency selects answers with the lowest frequencies. As the exemplar prompts are produced with answers with the highest frequencies, the Max Frequency strategy does not provide more information than exemplar prompts. In contrast, the Min Frequency strategy chooses captions that can provide some information not in the QA pairs, providing a performance boost.



\subsection{Ablation Study on Prompt Design} \label{exp:prompt_design}
We have two options to construct LLM's prompt. The first option is to append a syntheic QA pair after the caption that the QA pair is generated from. This can be described as CQA-CQA-CQA, where C, Q, A stand for caption, synthetic question, and synthetic answer respectively. Alternatively, we can present all captions at once, followed by all question-answer pairs, which we denote as CCC-QAQAQA. Experimentally (Table \ref{tab:cqa}), the second design performs significantly better than the first. We hypothesize that the first design may induce the LLM to read only one caption before answering, since in the prompt this caption contains all the information needed for the question. While it is hard to pinpoint the actual mechanism, the results highlight the importance of QA prompts and their positions.



 


\begin{figure*}[!htb]
		\vspace{-0.2em}
\centering
			         \begin{minipage}[c]{\linewidth}
				             \centering
				             \label{exp:aba_per}
 \includegraphics[width=0.98\linewidth]{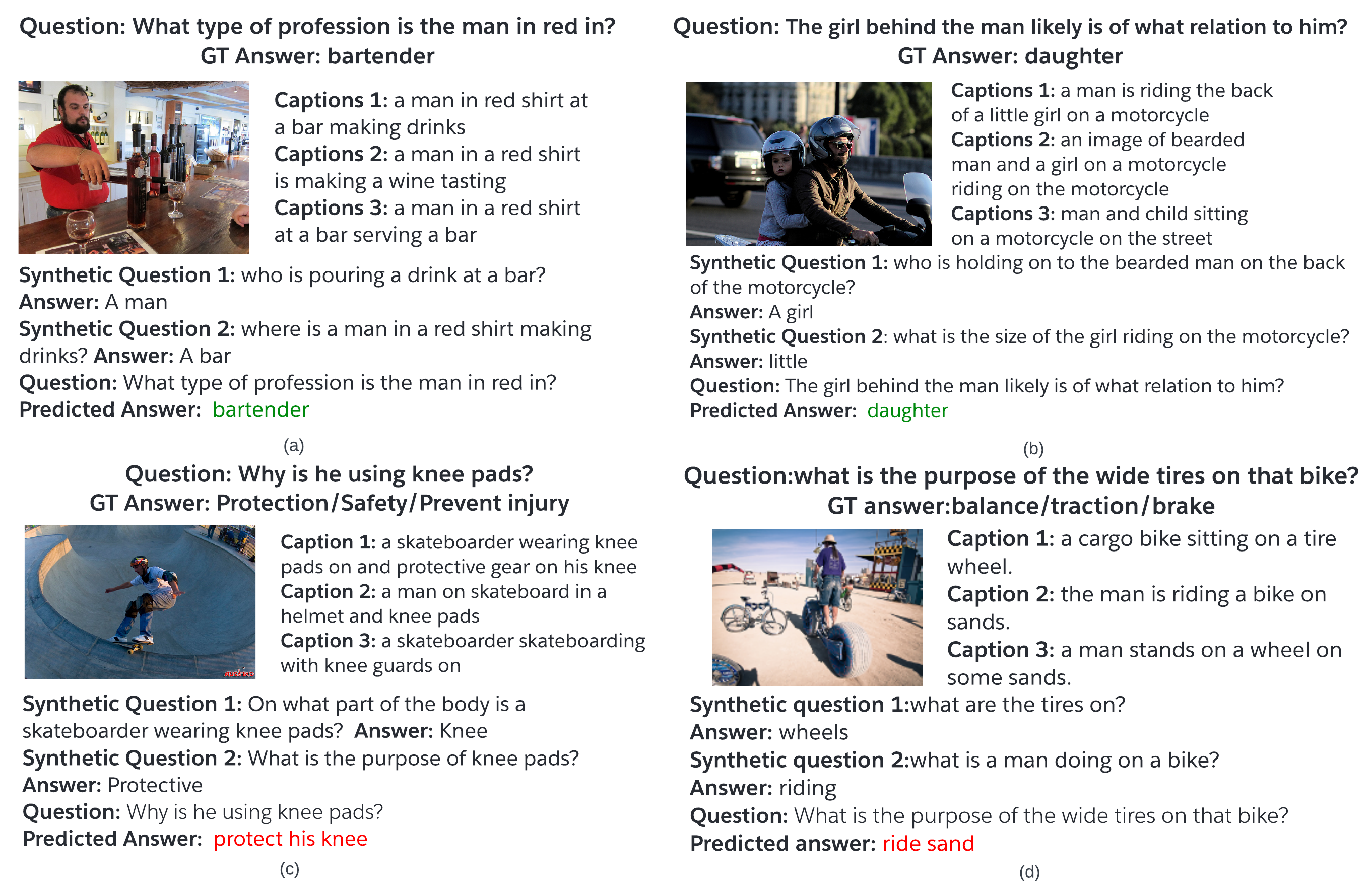}
				         \end{minipage}
			        
		
		\vspace{-0.8em}
	\caption{Example predictions made by Img2LLM.  Specifically, (a) and (b) are successful cases, while (c) and (d) are failure cases. See more examples at Appendix A.5.}
		\vspace{-0.8em}
  \label{fig:fail}
\end{figure*}

\subsection{Examples and Failure Case Analysis} \label{sec:quality_example}
In Figure \ref{fig:fail}, we show four examples of caption and exemplar prompts and the predictions, including cases of success and failure.  In Figure \ref{fig:fail}(a), the captions and the synthetic QA pairs provide the information that a man is making drinks at a bar. The LLM draws on background knowledge and correctly infers that his job is bartender. In Figure \ref{fig:fail}(c), while the prediction is understandable (even if not strictly grammatical), the LLM is unable to make inferences based on qualitative physics and predict the right answer. These results highlight the importance to apply appropriate commonsense knowledge in open-ended VQA.

 
				             
			        
		


%% file: tables/table3-qselect.tex
\begin{table}
\centering
\caption{Effect of question selection strategies.}
\vspace{-0.5em}
\label{tab:question_selection}
\begin{tabular}{cccc}
\hline 
   & &       OK-VQA & VQAv2\\
\hline
 \multicolumn{2}{c}{PICa\textsubscript{175B}} & 17.7 &- \\
 \hline
\multirow{1}{*}{Agnostic}& Random & 35.9 & 52.9 \\
  \hline
 \multirow{2}{*}{Template}& Random & 40.2 & 53.0 \\
  & Max Freq. & 41.5 & 55.8 \\
 \hline
\multirow{2}{*}{Neural}
 & Random  & 40.5  & 57.0\\
  & Max Freq. & \bf41.8 &\bf59.5  \\
                                  \hline
\end{tabular}
\vspace{-1.2em}
\end{table}

%% file: tables/table4.tex
\begin{table*}[!htb]
\centering
\vspace{-0.5em}
\caption{The experimental results on QA pairs generated from different captions. The results are run with OPT 30B.}
\label{tab:answer}
\vspace{-0.5em}
\setlength\tabcolsep{3.5pt}    
\begin{tabular}{cccccccccc}
\hline 
\multirow{3}{*}{\begin{tabular}[c]{@{}c@{}}Exemplar Prompts\\ Generation Source\end{tabular}}&       \multicolumn{3}{c}{OK-VQA}&  \multicolumn{3}{c}{VQAv2 val}\\
& VQA  & Answer  & Answer  & VQA  & Answer & Answer\\    
& Score & Noise Rate & Hit Rate & Score & Noise Rate & Hit Rate\\
                                  \hline
Caption from Complete Image & 39.8&0.018&0.480 & 57.1 &0.0290&0.725\\
Question-relevant Caption & 40.6 &0.022&\bf0.581& 58.1  &0.0303&\bf0.821\\
Question-relevant Caption with Filter & \bf41.8&\bf0.025&0.566& \bf59.5& \bf0.0313&0.804\\
                                  \hline
\end{tabular}
\vspace{-0.3em}
\end{table*}

%% file: 5-limitation.tex

One limitation of the proposed approach is that 
generating image captions and question-answer pairs incurs extra inference overhead.
On an 8$\times$A100 machine, our current implementation brings about 24.4\% additional computational time on top of the inference time of 175B OPT. We note that further reduction of  the overhead can be obtained by shortening the prompt, trading  accuracy for speed. Notably, our method avoids expensive end-to-end multimodal representation alignment, which took more than 500K TPU hours in the case of Flamingo. 

%



%% file: 6-conc.tex
In this paper, we propose \sysname, a plug-and-play module designed to exploit the knowledge and reasoning power of large language models (LLMs) off-the-shelf for zero-shot VQA tasks. Concretely,
\sysname{} provides visual information and task guidance to LLMs in the format of easily-digestible prompts.
This eliminates the requirement for the expensive end-to-end vision-language alignment, increasing model deployment flexibility while decreasing model deployment cost. The experiments show that \sysname{} enables different LLMs to achieve comparable or even superior zero-shot VQA performance to other methods that require costly end-to-end training. 

%% file: 7-appendix.tex



\appendix
\onecolumn
\section{Appendix}
\input{7-repro.tex}
\subsection{Broader Impact Statement}
We acknowledge that while the Img2LLM achieves comparable or superior performance to other zero-shot VQA methods, it has not reduced the inherent bias of these systems. Social-economic biases based on gender, age, race, and ethnicity exist in the datasets, LLMs, and VQA systems presented in this paper, including Img2LLM. Future work could assess the magnitude of this bias and mitigate its impact.

\subsection{Details about Question-Relevant Caption Generation}\label{app:gradcam}

Concretely,  we denote features of image patches extracted by ITE as $f_v^i \in \displaystyle \R^{K\times D^i_v}$ and question features as $f_q^i \in \displaystyle \R^{L \times D_q^i}$, where $i$ is the number of the layer of ITE, $K$ is the number of images patches, $L$ is the number of token in the given question, $D^i_v$ is the dimension of patch feature in the $i$-{th} layer of ITE network and $D^i_q$ is the dimension of textual feature in the $i$-{th} layer of ITE network.  For cross-attention head in $i$-{th} layer, the cross-attention scores $W^i$ between each image patch and each token in question can be calculated directly as
\begin{equation}
\vspace{-0.1em}
\footnotesize
    W^i = \operatorname{softmax}\left(\frac{f_q^i W_Q^i {W_K^i}^\top {f_v^i}^\top}{\sqrt{D_q^i}}\right).
    \label{eq:cross-a}
    \vspace{-0.1em}
\end{equation}

where $W_Q^i \in \displaystyle \R^{D_q^i \times D_q^i}$ is the query head and $W_K^i\in \displaystyle \R^{D_v^i \times D_q^i}$ is the key head in the $i$-{th} layer of ITE network. With Equation \ref{eq:cross-a}, we obtain a cross-attention matrix $W^i \in \displaystyle \R^{L\times K}$, where each row is the cross-attention scores of each token in the question over all image patches. Specifically, the attention matrix $W^i$ can be regarded as the patch importance for ITE to calculate the similarity of whole image and question, but it still contains redundancy that contributes only a minor performance loss \cite{bian-etal-2021-attention}, indicating that some patches are uninformative. In order to find these less relevant image patches, we follwing GradCAM and compute the derivative of the cross-attention score from ITE function $\text{ sim}(v, q)$, \emph{i.e.}, $\partial \text{ sim}(v, q) / \partial W$, and multiplying its gradient matrix with the cross-attention scores element-wisely. 
The relevance of the $k$\textsuperscript{th} image patch with the question, $r_k^i$, can be computed as the average over $H$ attention heads and the sum over $L$ textual tokens:
\begin{equation}
\vspace{-0.1em}
\footnotesize
    r_k^i = \frac{1}{H} \sum^{L}_{l=1} \sum^{H}_{h=1} \min\left(0,\frac{\partial  \text{ sim}(v, q)}{\partial W_{lk}^{ih}}\right) W_{lk}^{ih},
\vspace{-0.1em}
\end{equation}
where $h$ is the index of attention heads and $i$ is the layer index of ITE.

\subsection{Experimental Results of Supervised Learning Methods in A-OKVQA}\label{app:aokvqa}

We show the experimental comparisons between our method and supervised model on A-OKVQA dataset \cite{schwenk2022okvqa} as Table \ref{tab:sensitive} shows. We can observe that our method outperform almost all supervised model with smaller size language model. This strongly support our method's effectiveness in leveraging reasoning power of large language models.

\begin{table*}[!htb]
\centering
\caption{The experimental comparisons with models trained in A-OKVQA training dataset.}
\begin{tabular}{ccc}
\hline
Methods & \multicolumn{2}{c}{A-OKVQA} \\
                          & Val                      & Test \\
                          \hline
                           \multicolumn{3}{c} {\emph{ Models Fine-Tuned in A-OKVQA Training Set }} \\  
Pythia \cite{jiang2018pythia}                & 25.2                     & 21.9 \\
ViLBERT \cite{lu2019vilbert}                   & 30.6                     & 25.9 \\
LXMERT \cite{tan2019lxmert}                    & 30.7                     & 25.9 \\
KRISP    \cite{marino2021krisp}                 & 33.7                     & 27.1 \\
GPV-2 \cite{kamath2022webly}& \bf48.6 & \bf40.7 \\
\hline
 \multicolumn{3}{c} {\emph{ Zero-Shot Evaluation with Plug-in Frozen Large Language Model}}  \\
  Ours\textsubscript{6.7B}                                                    & 33.3& 32.2 \\
                           Ours\textsubscript{13B}                                                    & 33.3&33.0 \\
Ours\textsubscript{30B}                                                    & 36.9 &36.0 \\
Ours\textsubscript{66B}                                                    & 38.7 &38.2\\ 

                                   Ours\textsubscript{175B}                                                     &{42.9} & \textbf{40.7}  \\
                                   \hline
\end{tabular}
\end{table*}

\subsection{Template-Based Question Design} \label{appsec:rule_question}

We design question templates for each part of speech type of answers as Table \ref{tab:template} shows. 
\begin{table*}[!htb]
\centering
\caption{The question templates for answers with different part of speech.}
\label{tab:template}
\begin{tabular}{c|c}
\hline
Part of Speech of Answer
          & Question Templates           \\
          \hline
Noun      & \begin{tabular}[c]{@{}c@{}}What item is this in this picture?\\ What item is that in this picture?\end{tabular}                                                                                                                                                            \\
 \hline
Verb      & \begin{tabular}[c]{@{}c@{}}What action is being done in this picture?\\ Why is this item doing in this picture?\\ Which action is being taken in this picture?\\ What action is item doing in this picture?\\ What action is item performing in this picture?\end{tabular} \\
 \hline
Adjective & \begin{tabular}[c]{@{}c@{}}How to describe one item in this picture?\\ What is item's ADJ TYPE in this picture?\\ What is the ADJ TYPE in this picture?\end{tabular}                                                                                                       \\
 \hline
Num       & How many things in this picture?   \\
 \hline
\end{tabular}
\end{table*}

\begin{table*}[!htb]
\centering
\caption{The experimental results of using different number of captions and QA pairs as prompts. The experiments are run on OK-VQA with OPT 30B. }
\label{tab:sensitive}
\begin{tabular}{cccccccc}
\hline 
                                                                            \diagbox{QA Pairs}{Caption}&       0 & 10 & 20 &30 & 40 & 50\\
                                  \hline
 0  & 3.3 & 19.6&22.7&23.4&24.0 &24.8\\
 10 &40.9 &41.6 &42.1 &42.1&41.9&42.2\\
 20&41.2 & 41.3& 41.3&41.7 &42.2 &42.0\\
 30 & 41.0 &41.0 &41.7 &41.8&41.6&41.5\\
 40 &40.3 &40.7 &40.6 &40.3 & 40.3 &41.1\\
 50 & 40.6&40.6 & 40.7&40.9&40.6 &41.1\\
 
                                  \hline
\end{tabular}
\end{table*}

\subsection{Sensitive Analysis} \label{sec:sensitivity}

We evaluate the sensitive analysis about the QA pairs and number of captions in prompt for LLM as Table \ref{tab:sensitive} shows. We can observe that the differences in QA scores on OK-VQA dataset are not higher than 1 as long as QA pairs in prompts. The results demonstrate the performance of  our method is robust with different numbers of QA pairs and captions.


\begin{table}[!htb]
\centering
\caption{The experimental results of using different number of patches to generate question-relevant captions. The experiments are run on OK-VQA with OPT 30B. }
\begin{tabular}{ccccc}
\hline 
                                                                            Patch\_num&        10 & 20  & 40 & Full\\
                                  \hline
   & 41.2&41.8 & 41.6 & 39.8\\

                                  \hline
\end{tabular}
\end{table}


\begin{table}[!htb]
\centering
\caption{The experimental results of generating different number of question-relevant captions. The experiments are run on OK-VQA with OPT 30B. }
\begin{tabular}{cccccc}
\hline 
                                                                            Caption\_num&   PICa    & 10 & 30  & 50 & 100\\
                                  \hline
 & 17.7 & 38.3 & 40.9 & 41.4&41.8\\

                                  \hline
\end{tabular}
\end{table}

\subsection{Examples} \label{appsec:example}
\input{example_appendix.tex}

%% file: 7-repro.tex
\subsection{Reproducibility Statement }
We acknowledge the importance of reproducibility for research work and try whatever we can to ensure the reproducibility of our work.  As for the implementation of our method, details such as hyperparameters are provided in Section~4.1 in the main paper. We will publicly release all codes after the acceptance of this paper.  

%% file: example_appendix.tex
\begin{figure*}[!htb]
		\vspace{-0.5em}
\centering
		\subfloat[]{      
			         \begin{minipage}[c]{\linewidth}
				             \centering
 \includegraphics[height=1.6in]{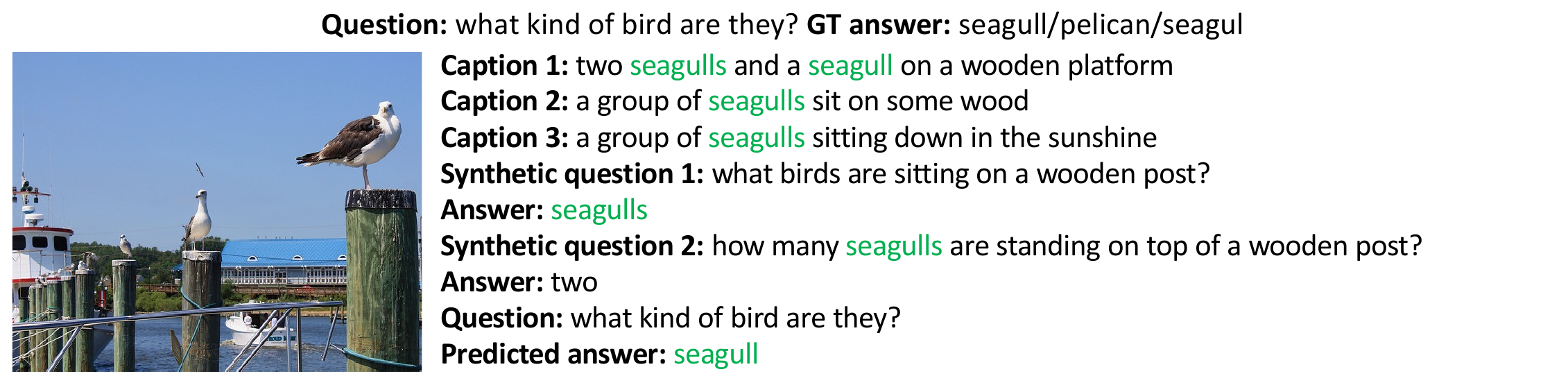}
				         \end{minipage}
			     }\vspace{-1.em}
			     \\
			     \subfloat[]{      
			         \begin{minipage}[c]{\linewidth}
				             \centering
 \includegraphics[height=1.6in]{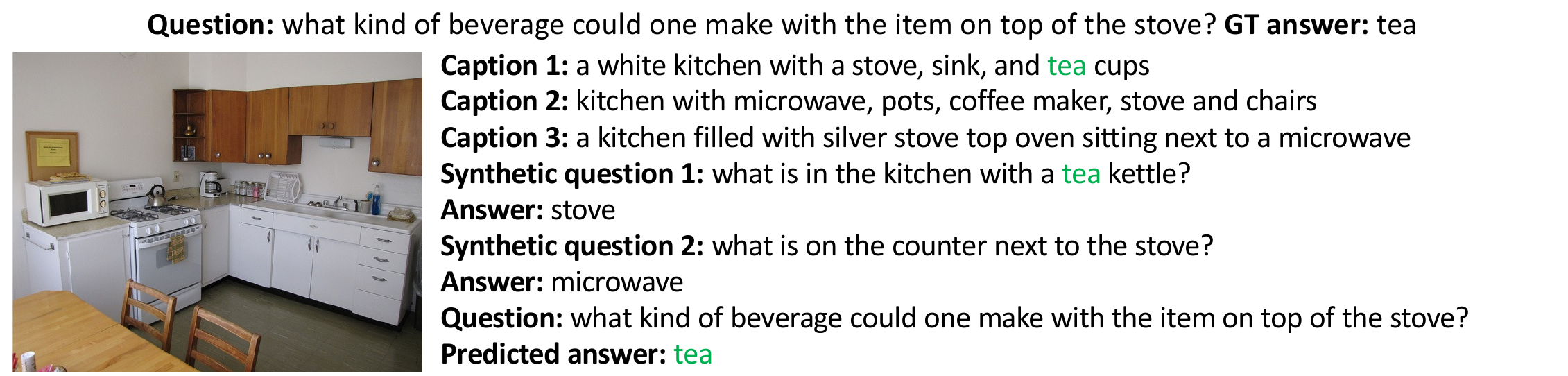}
				         \end{minipage}
			     }
					     \\
		\subfloat[]{      
			\begin{minipage}[c]{\linewidth}
				\centering
				\includegraphics[height=1.6in]{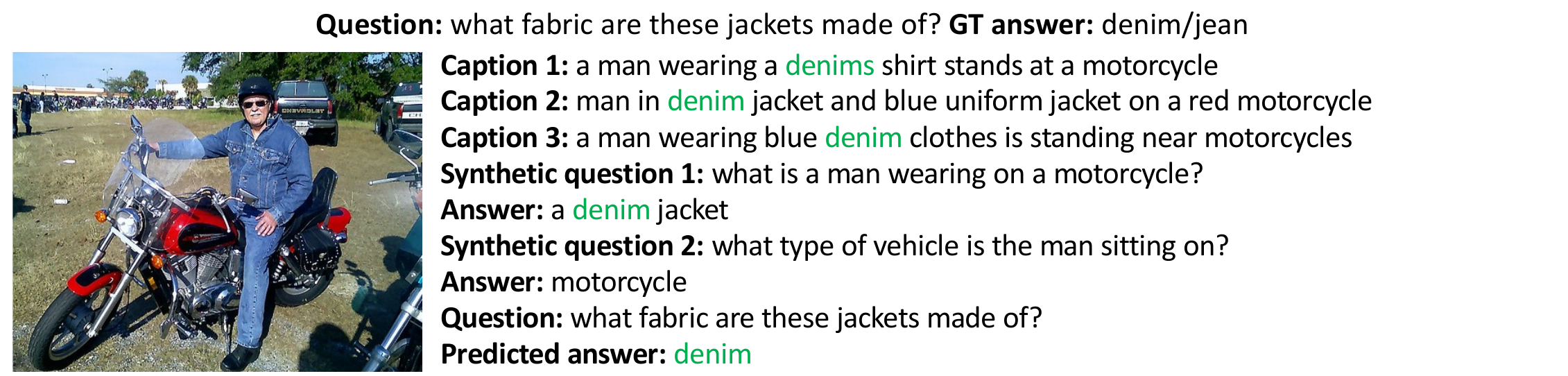}
			\end{minipage}
		}
					     \\
		\subfloat[]{      
			\begin{minipage}[c]{\linewidth}
				\centering
				\includegraphics[height=1.6in]{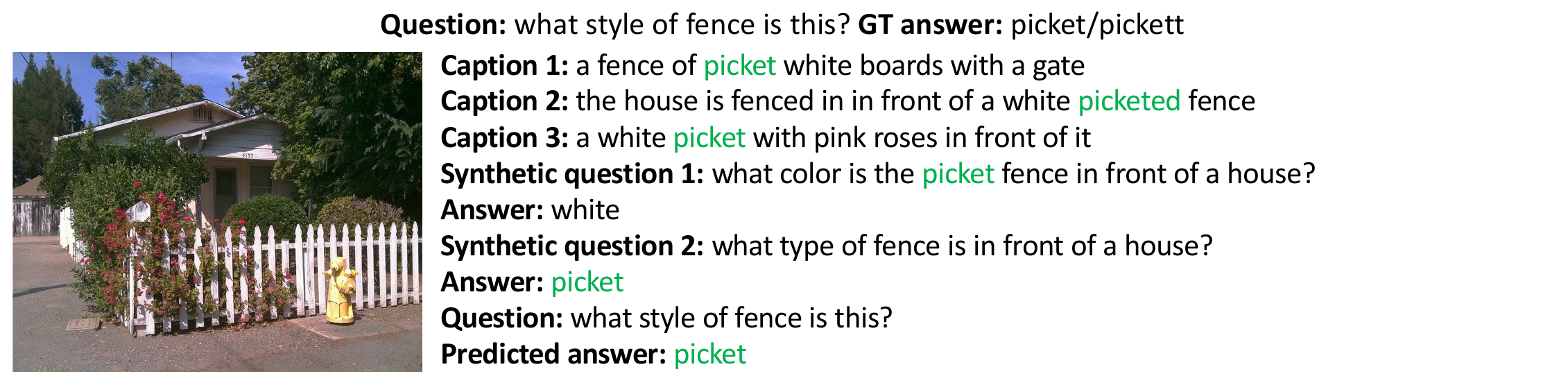}
			\end{minipage}
		}
				     \\
	\subfloat[]{      
		\begin{minipage}[c]{\linewidth}
			\centering
			\includegraphics[height=1.6in]{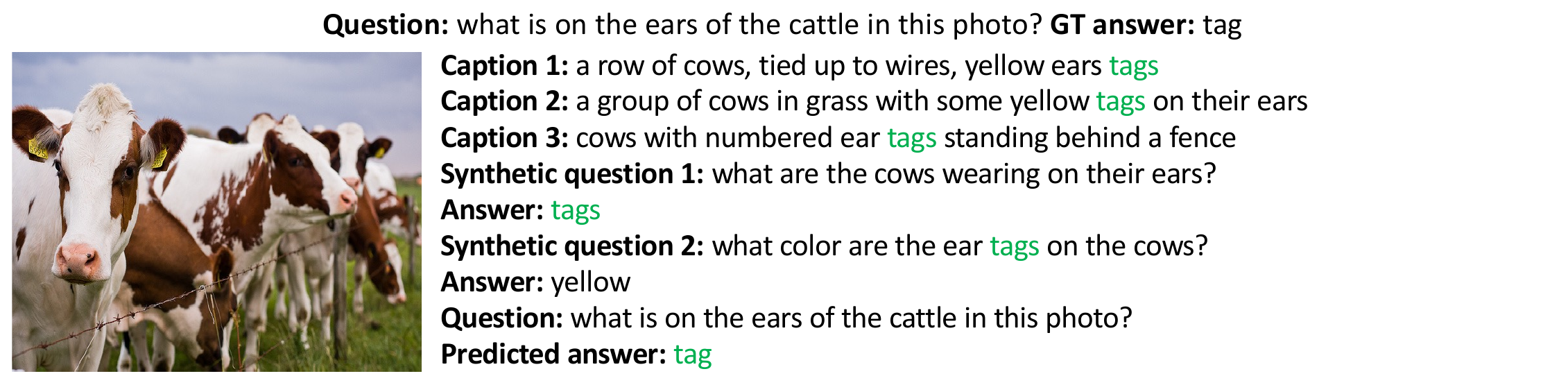}
		\end{minipage}
	}
		\vspace{-0.8em}

	\caption{{Success case analysis for OK-VQA. Green color indicates answer cues and correct prediction.} }
		\vspace{-1.em}
\end{figure*}

\begin{figure*}[!htb]
	\vspace{-0.5em}
	\centering
	\subfloat[]{      
		\begin{minipage}[c]{\linewidth}
			\centering
			\includegraphics[height=1.6in]{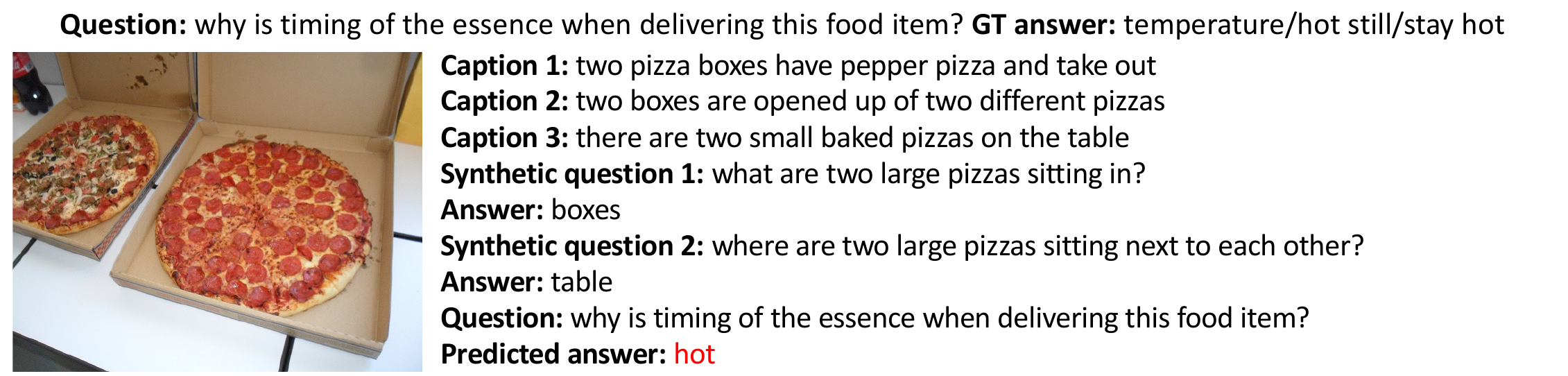}
		\end{minipage}
	}\vspace{-1.em}
	\\
	\subfloat[]{      
		\begin{minipage}[c]{\linewidth}
			\centering
			\includegraphics[height=1.6in]{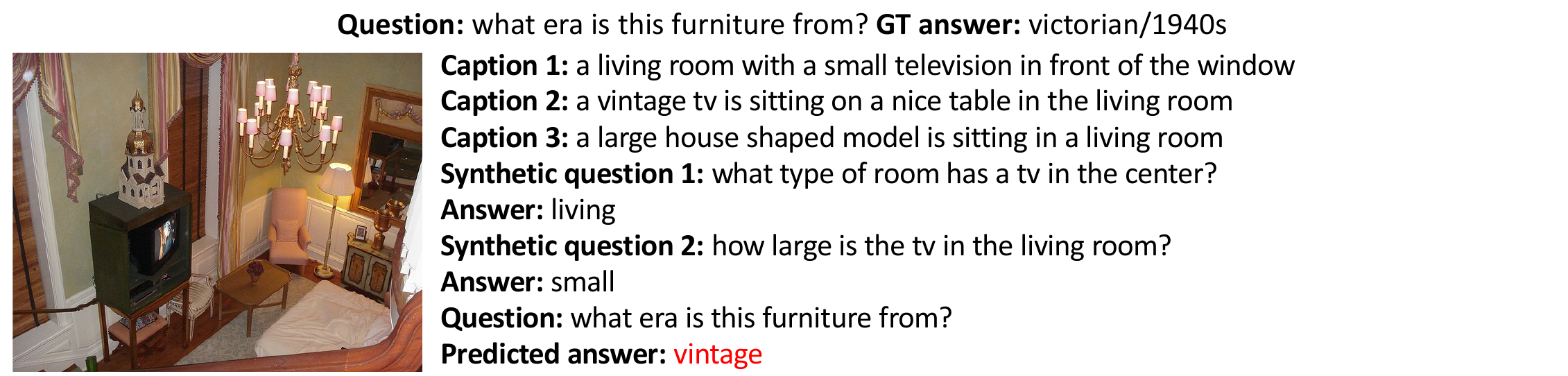}
		\end{minipage}
	}
	\\
	\subfloat[]{      
		\begin{minipage}[c]{\linewidth}
			\centering
			\includegraphics[height=1.6in]{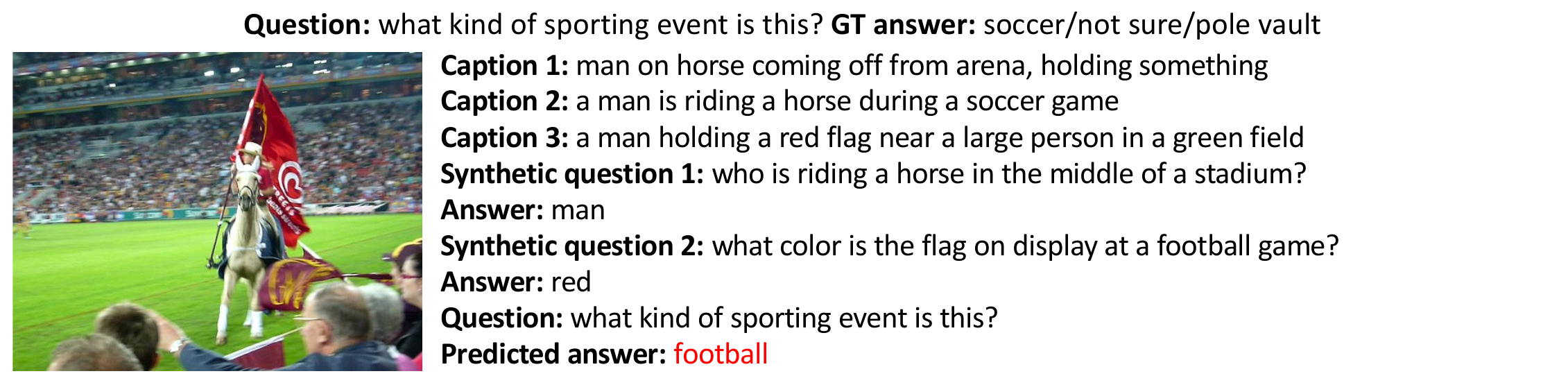}
		\end{minipage}
	}
	\\
	\subfloat[]{      
		\begin{minipage}[c]{\linewidth}
			\centering
			\includegraphics[height=1.6in]{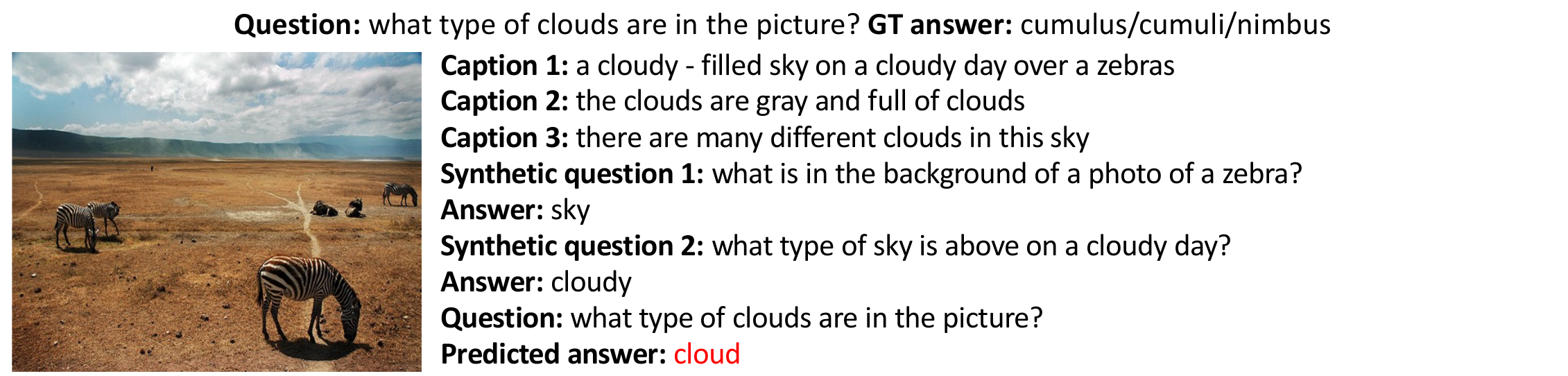}
		\end{minipage}
	}
	\\
	\subfloat[]{      
		\begin{minipage}[c]{\linewidth}
			\centering
			\includegraphics[height=1.6in]{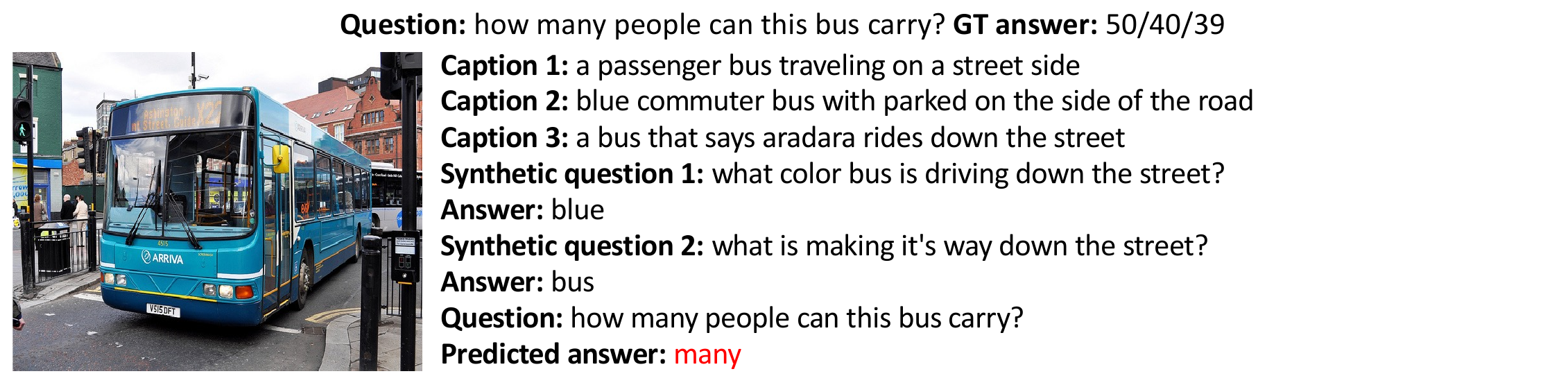}
		\end{minipage}
	}
	\vspace{-0.8em}
	
	\caption{{Failure case analysis for OK-VQA. Red color indicates incorrect prediction.} }
	\vspace{-1.em}
\end{figure*}

\begin{figure*}[!htb]
	\vspace{-0.5em}
	\centering
	\subfloat[]{      
		\begin{minipage}[c]{\linewidth}
			\centering
			\includegraphics[height=1.6in]{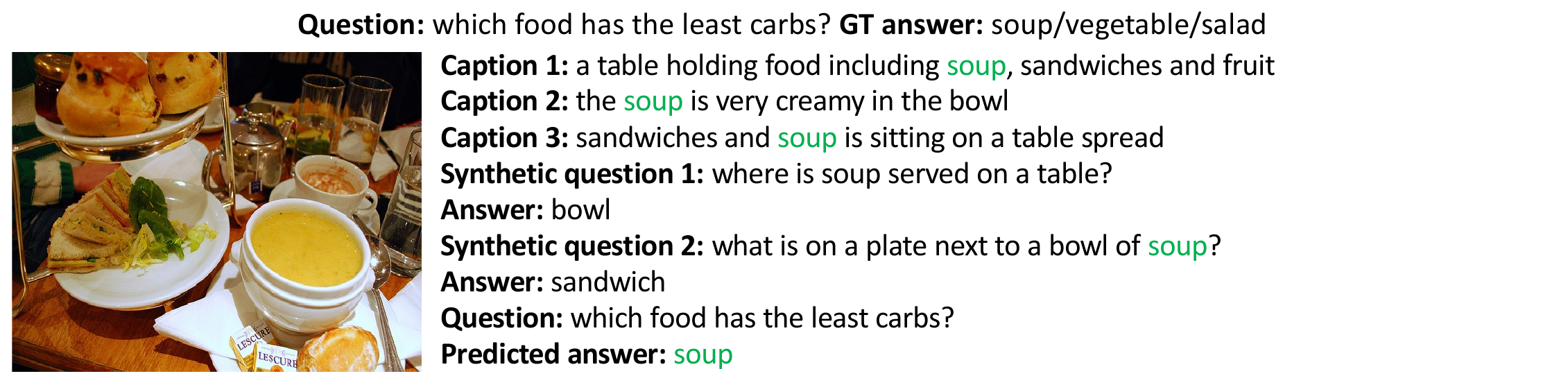}
		\end{minipage}
	}\vspace{-1.em}
	\\
	\subfloat[]{      
		\begin{minipage}[c]{\linewidth}
			\centering
			\includegraphics[height=1.6in]{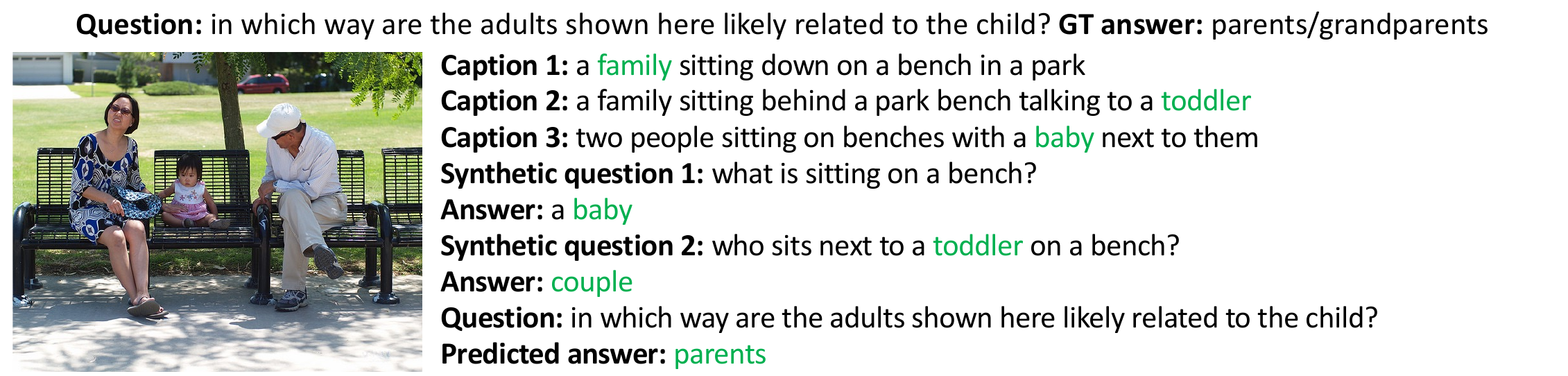}
		\end{minipage}
	}
	\\
	\subfloat[]{      
		\begin{minipage}[c]{\linewidth}
			\centering
			\includegraphics[height=1.6in]{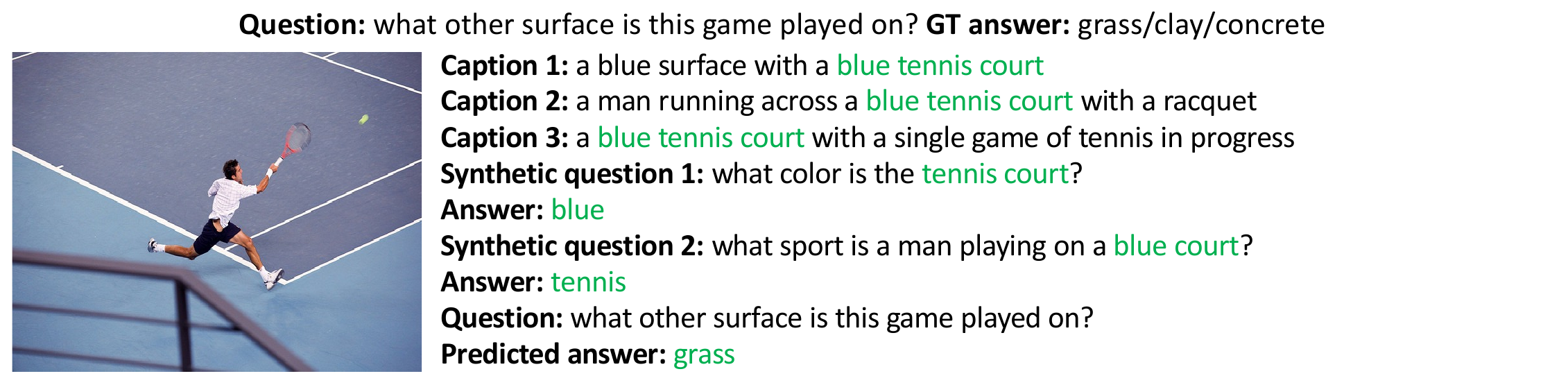}
		\end{minipage}
	}
	\\
	\subfloat[]{      
		\begin{minipage}[c]{\linewidth}
			\centering
			\includegraphics[height=1.6in]{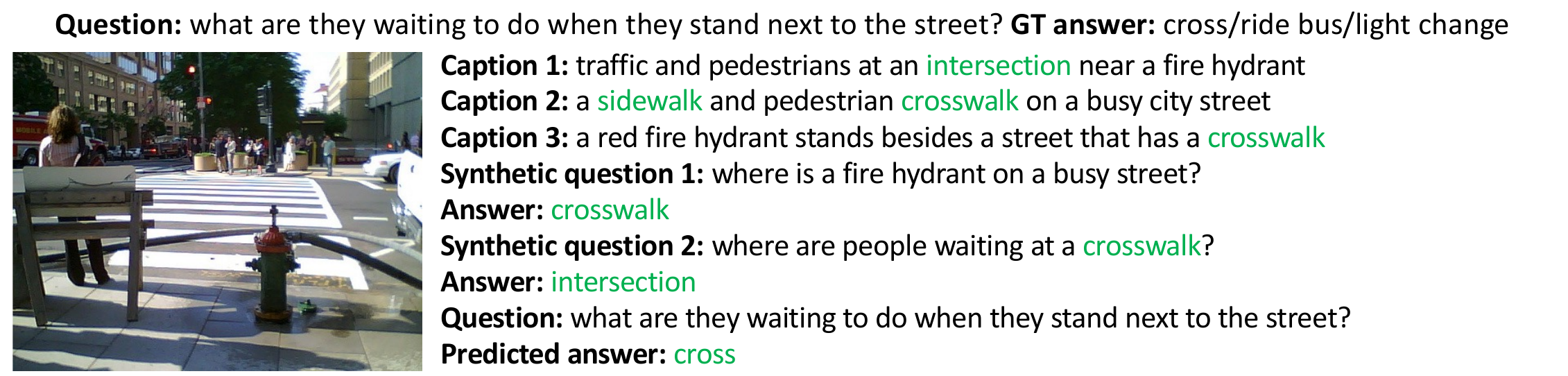}
		\end{minipage}
	}
	\\
	\subfloat[]{      
		\begin{minipage}[c]{\linewidth}
			\centering
			\includegraphics[height=1.6in]{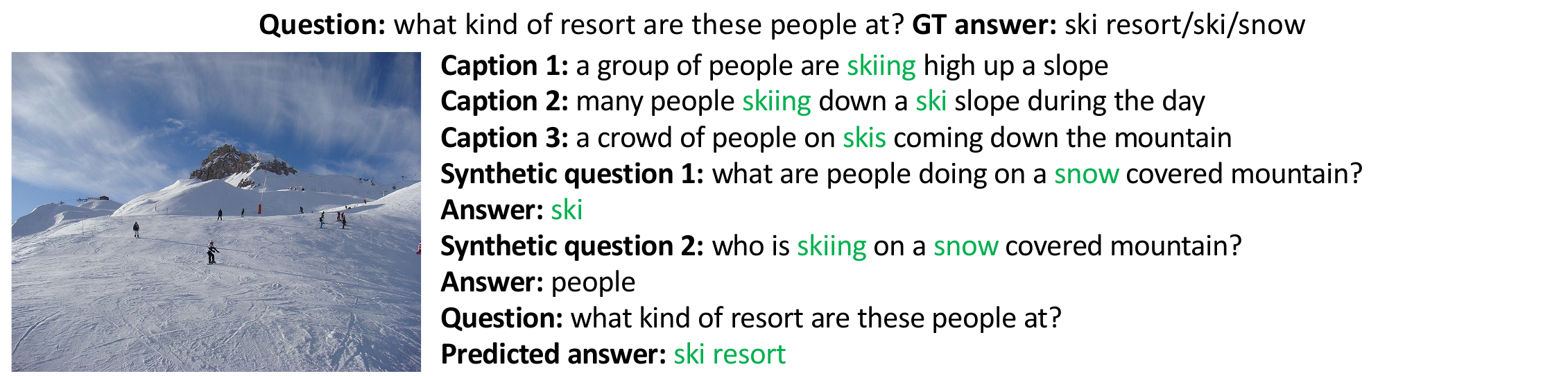}
		\end{minipage}
	}
	\vspace{-0.8em}
	
	\caption{{Success case analysis for A-OKVQA. Green color indicates answer cues and correct prediction.} }
	\vspace{-1.em}
\end{figure*}

\begin{figure*}[!htb]
	\vspace{-0.5em}
	\centering
	\subfloat[]{      
		\begin{minipage}[c]{\linewidth}
			\centering
			\includegraphics[height=1.6in]{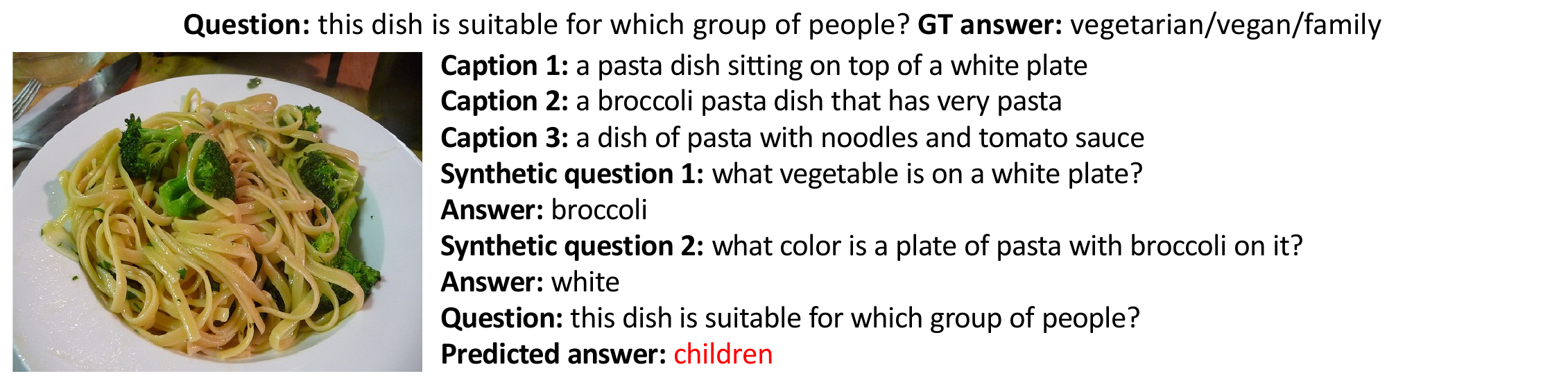}
		\end{minipage}
	}\vspace{-1.em}
	\\
	\subfloat[]{      
		\begin{minipage}[c]{\linewidth}
			\centering
			\includegraphics[height=1.6in]{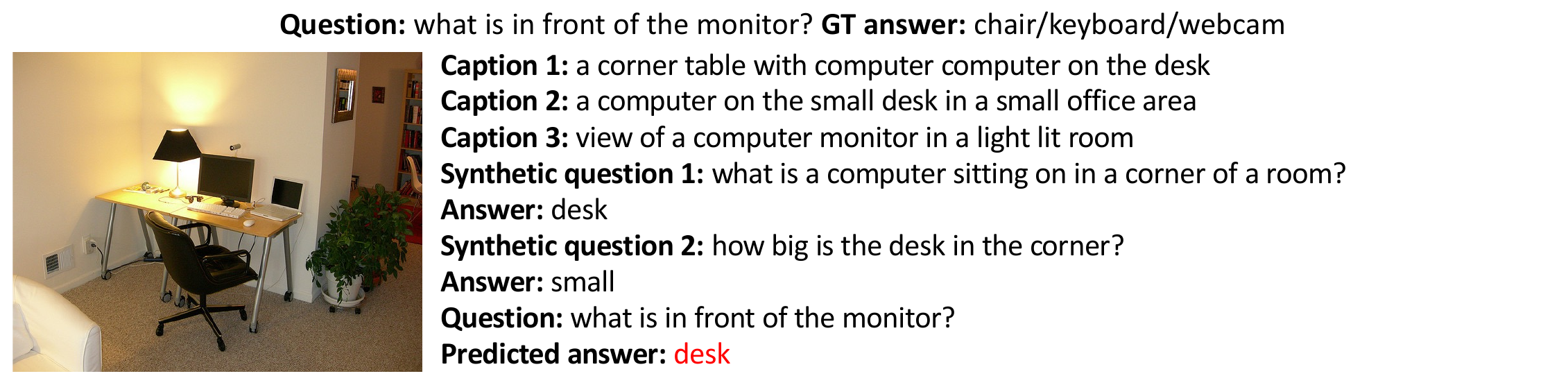}
		\end{minipage}
	}
	\\
	\subfloat[]{      
		\begin{minipage}[c]{\linewidth}
			\centering
			\includegraphics[height=1.6in]{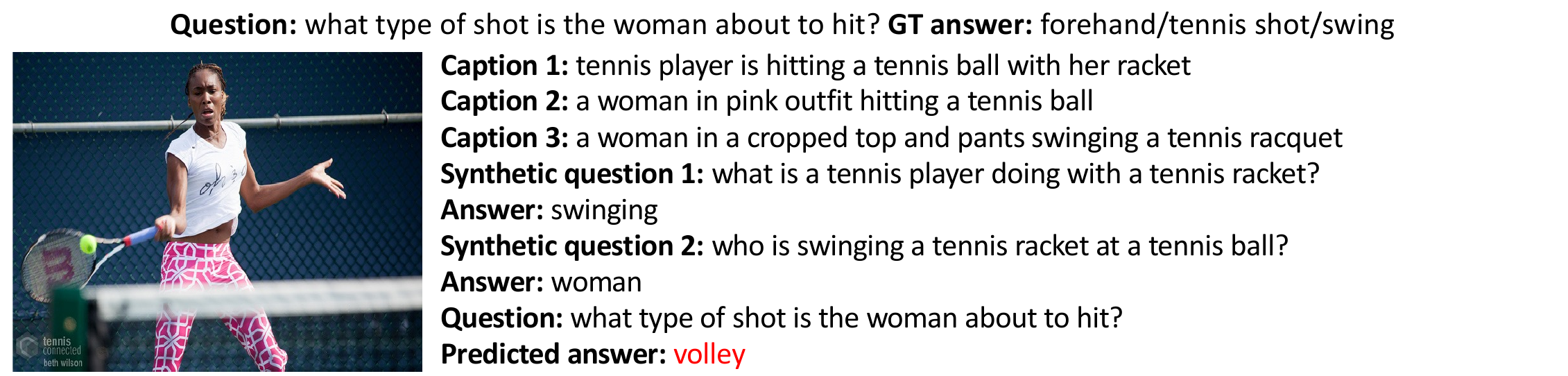}
		\end{minipage}
	}
	\\
	\subfloat[]{      
		\begin{minipage}[c]{\linewidth}
			\centering
			\includegraphics[height=1.6in]{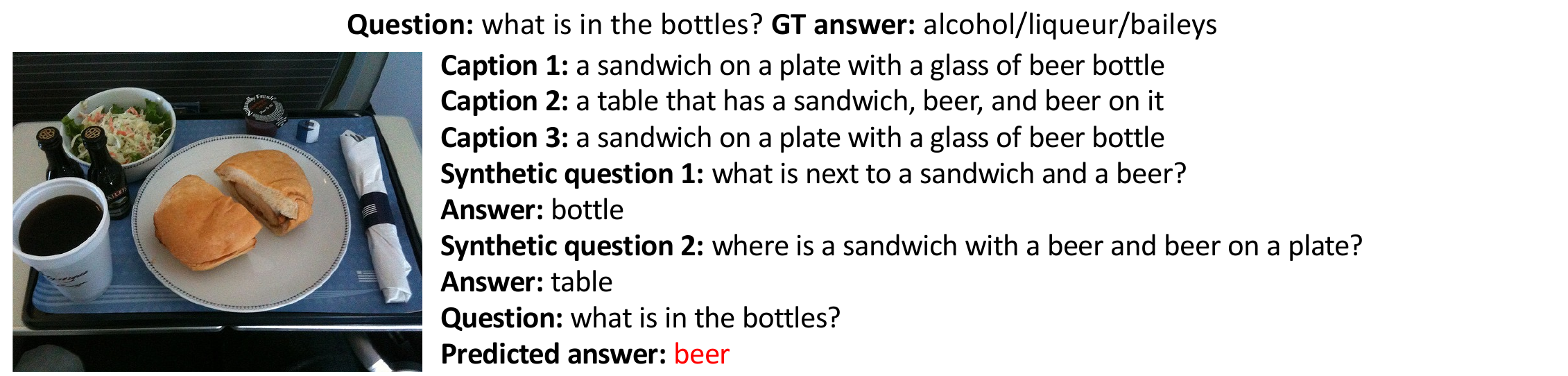}
		\end{minipage}
	}
	\\
	\subfloat[]{      
		\begin{minipage}[c]{\linewidth}
			\centering
			\includegraphics[height=1.6in]{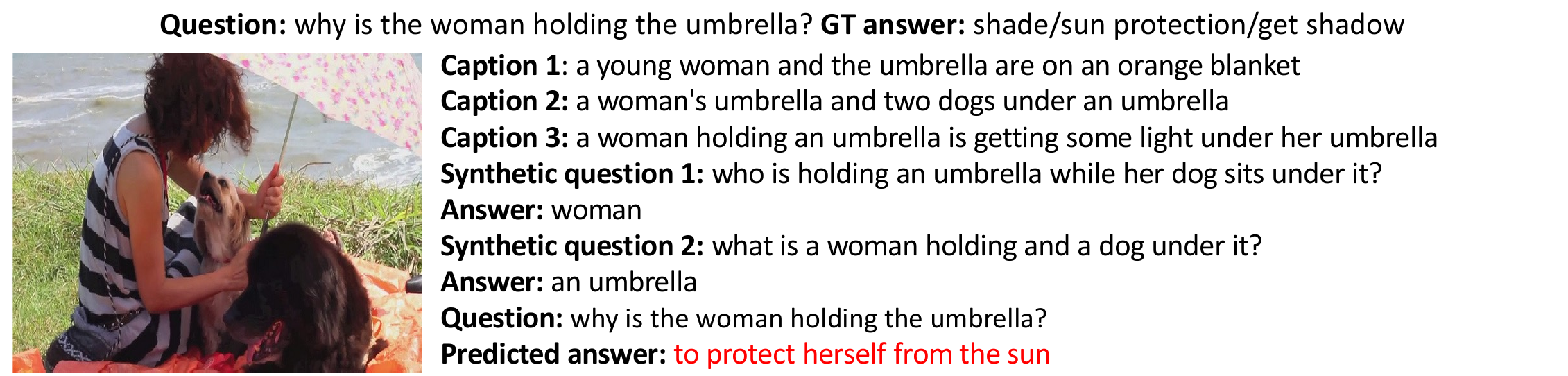}
		\end{minipage}
	}
	\vspace{-0.8em}
	
	\caption{{Failure case analysis for A-OKVQA. Red color indicates incorrect prediction.} }
	\vspace{-1.em}
\end{figure*}

\begin{figure*}[!htb]
	\vspace{-0.5em}
	\centering
	\subfloat[]{      
		\begin{minipage}[c]{\linewidth}
			\centering
			\includegraphics[height=1.6in]{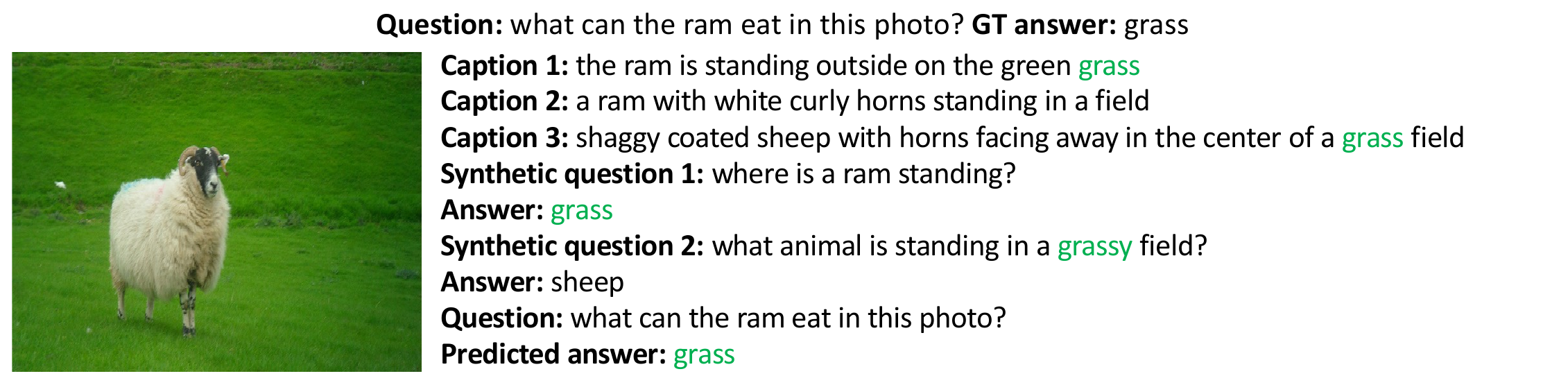}
		\end{minipage}
	}\vspace{-1.em}
	\\
	\subfloat[]{      
		\begin{minipage}[c]{\linewidth}
			\centering
			\includegraphics[height=1.6in]{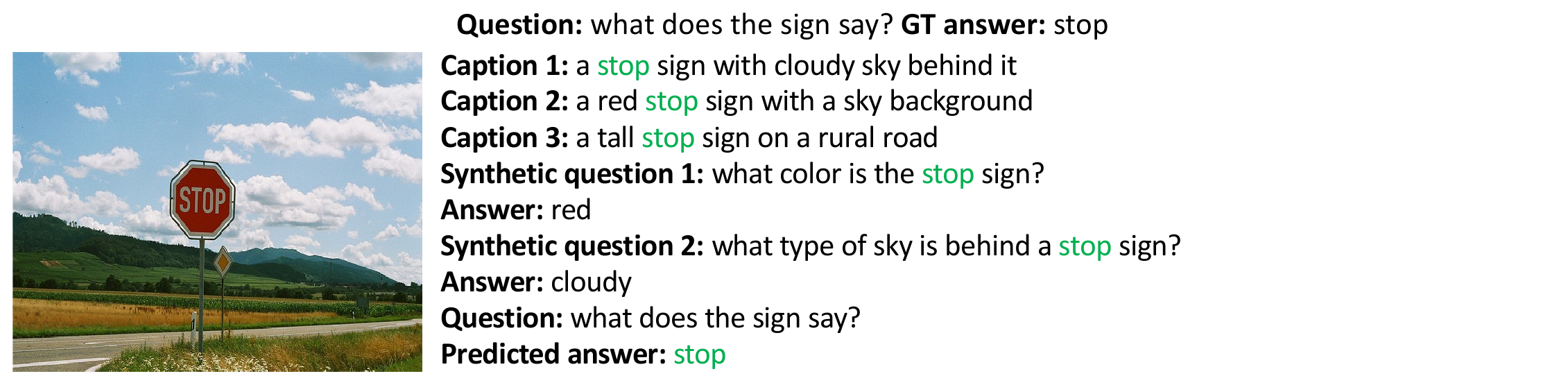}
		\end{minipage}
	}
	\\
	\subfloat[]{      
		\begin{minipage}[c]{\linewidth}
			\centering
			\includegraphics[height=1.6in]{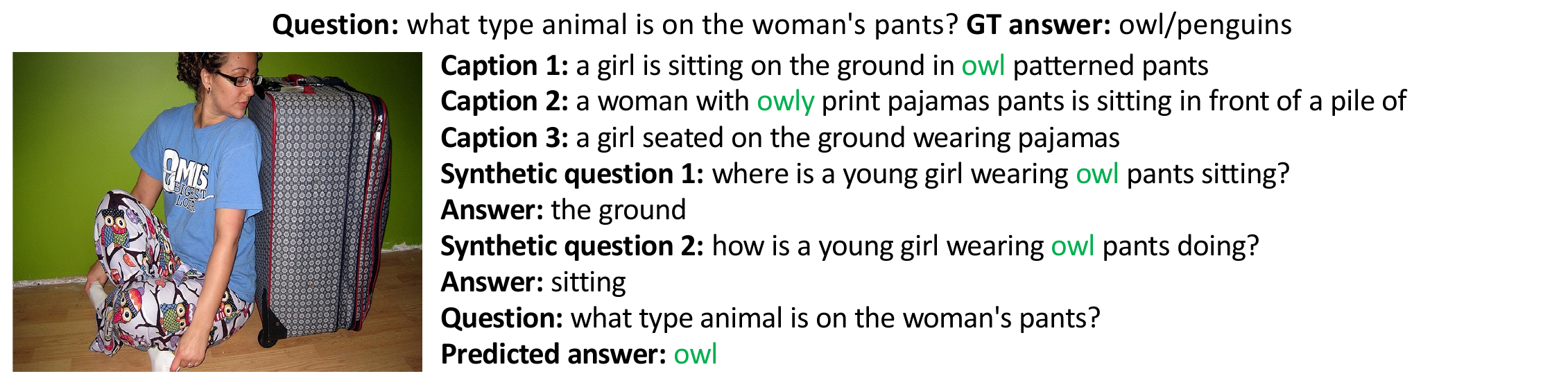}
		\end{minipage}
	}
	\\
	\subfloat[]{      
		\begin{minipage}[c]{\linewidth}
			\centering
			\includegraphics[height=1.6in]{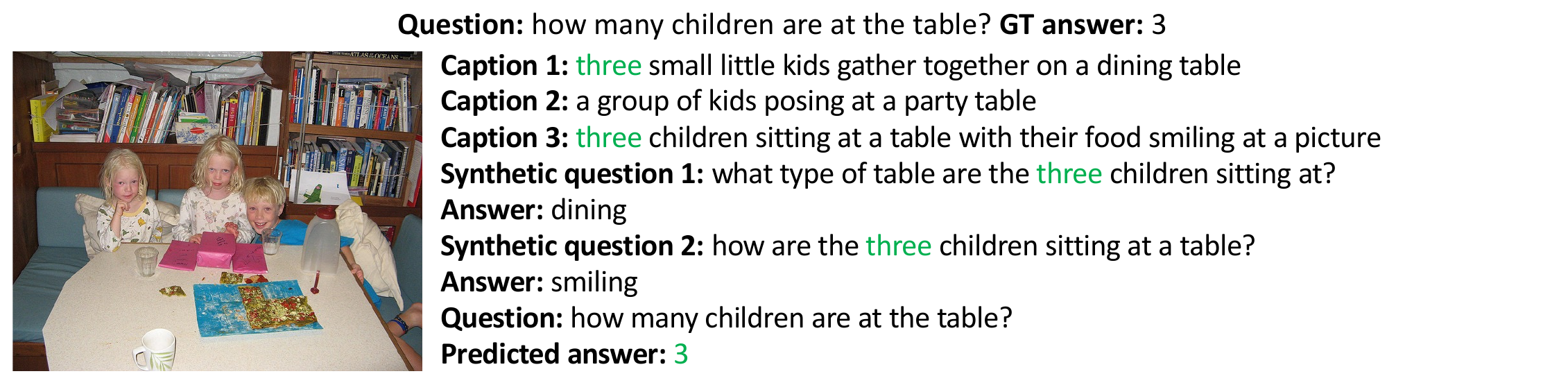}
		\end{minipage}
	}
	\\
	\subfloat[]{      
		\begin{minipage}[c]{\linewidth}
			\centering
			\includegraphics[height=1.6in]{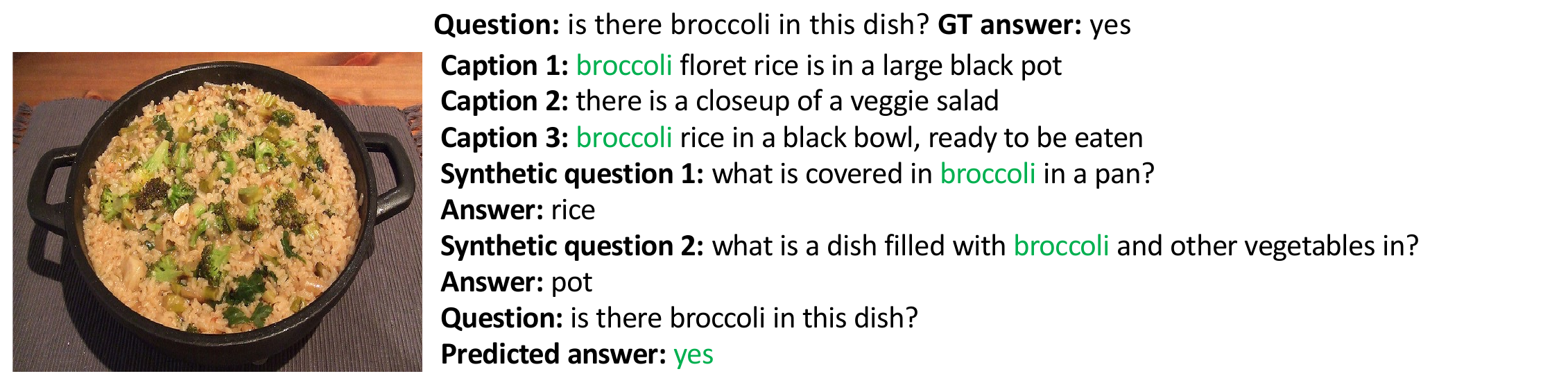}
		\end{minipage}
	}
	\vspace{-0.8em}
	
	\caption{{Success case analysis for VQAv2. Green color indicates answer cues and correct prediction.} }
	\vspace{-1.em}
\end{figure*}

\begin{figure*}[!htb]
	\vspace{-0.5em}
	\centering
	\subfloat[]{      
		\begin{minipage}[c]{\linewidth}
			\centering
			\includegraphics[height=1.6in]{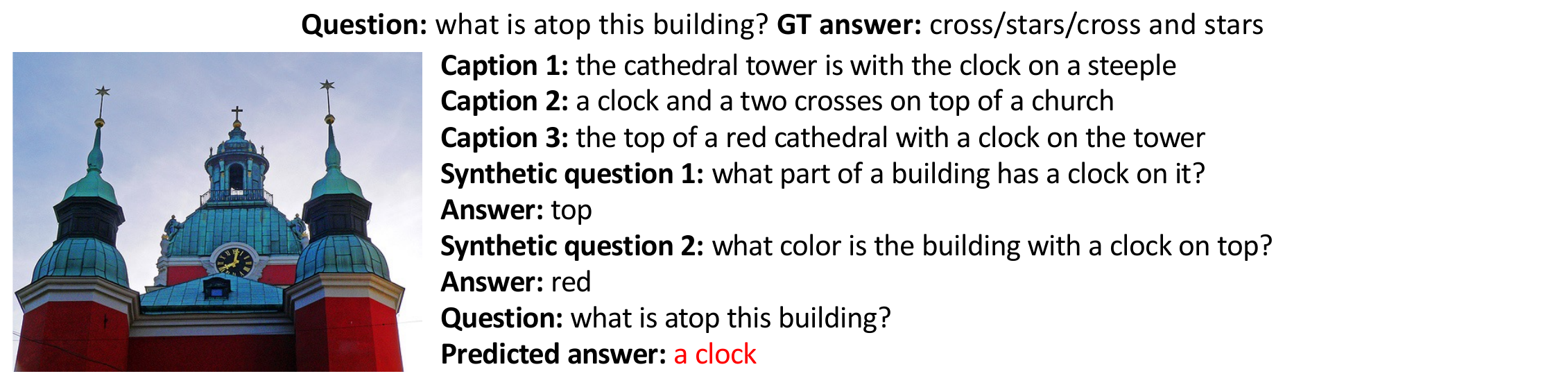}
		\end{minipage}
	}\vspace{-1.em}
	\\
	\subfloat[]{      
		\begin{minipage}[c]{\linewidth}
			\centering
			\includegraphics[height=1.6in]{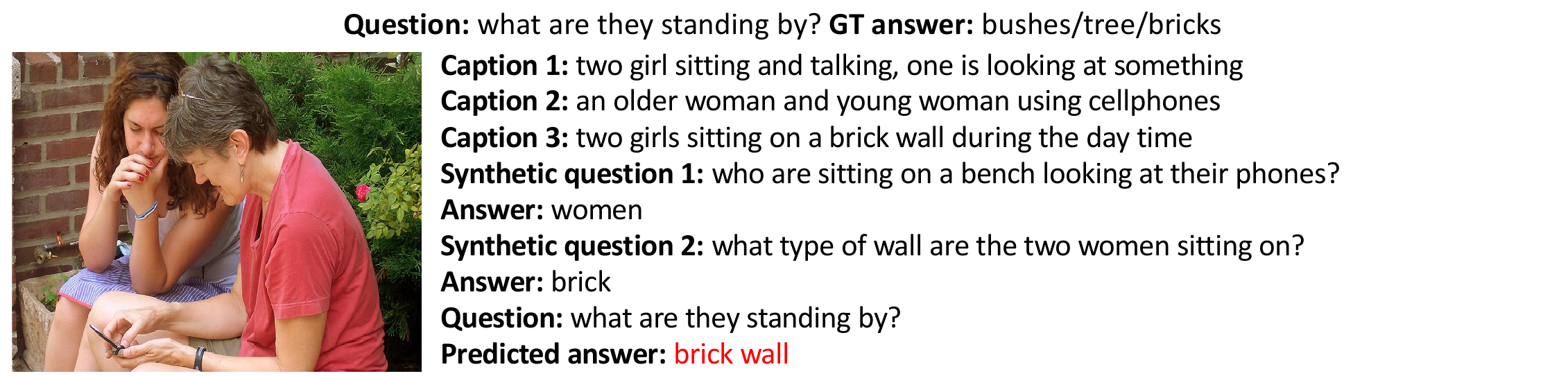}
		\end{minipage}
	}
	\\
	\subfloat[]{      
		\begin{minipage}[c]{\linewidth}
			\centering
			\includegraphics[height=1.6in]{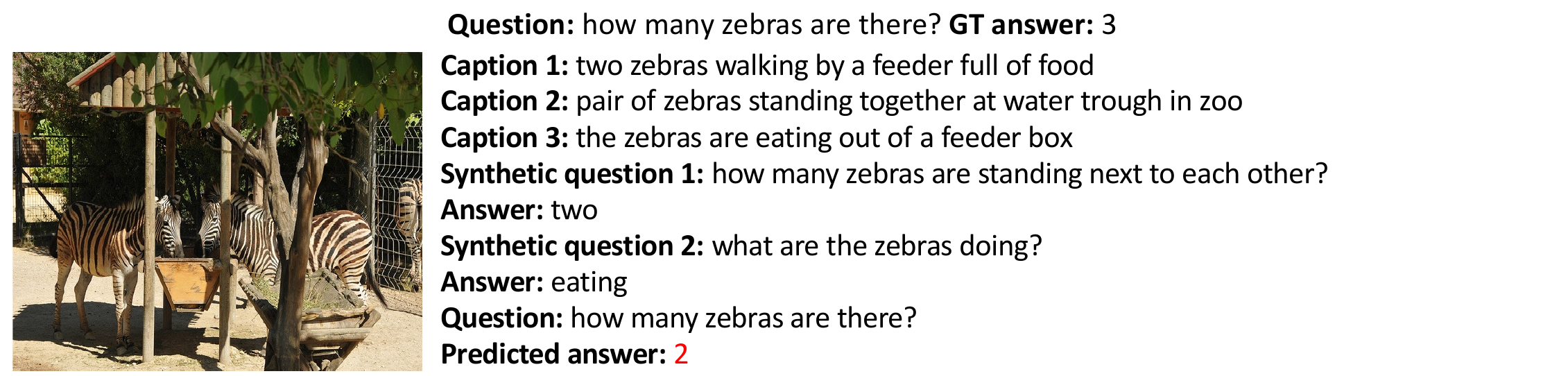}
		\end{minipage}
	}
	\\
	\subfloat[]{      
		\begin{minipage}[c]{\linewidth}
			\centering
			\includegraphics[height=1.6in]{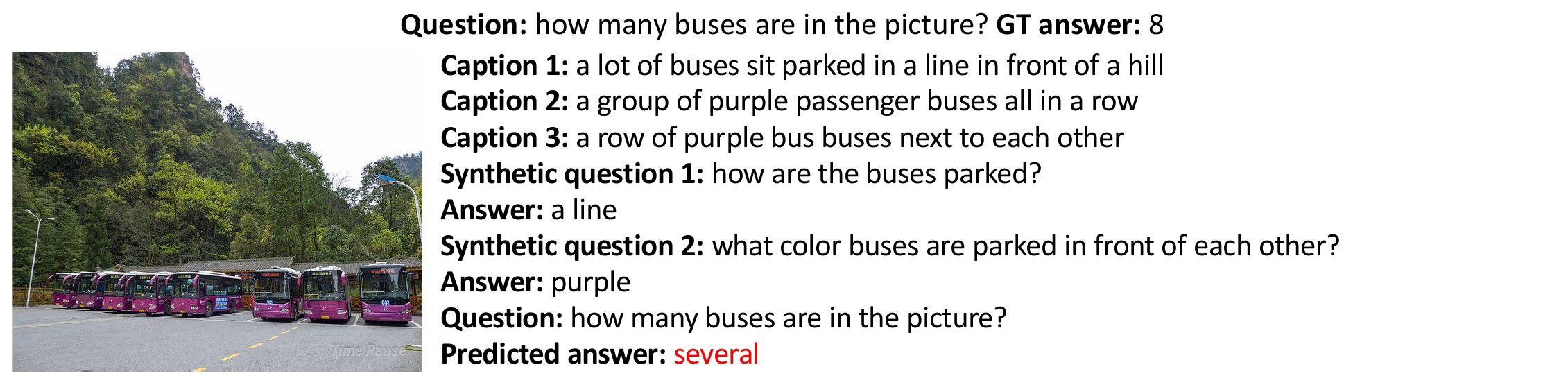}
		\end{minipage}
	}
	\\
	\subfloat[]{      
		\begin{minipage}[c]{\linewidth}
			\centering
			\includegraphics[height=1.6in]{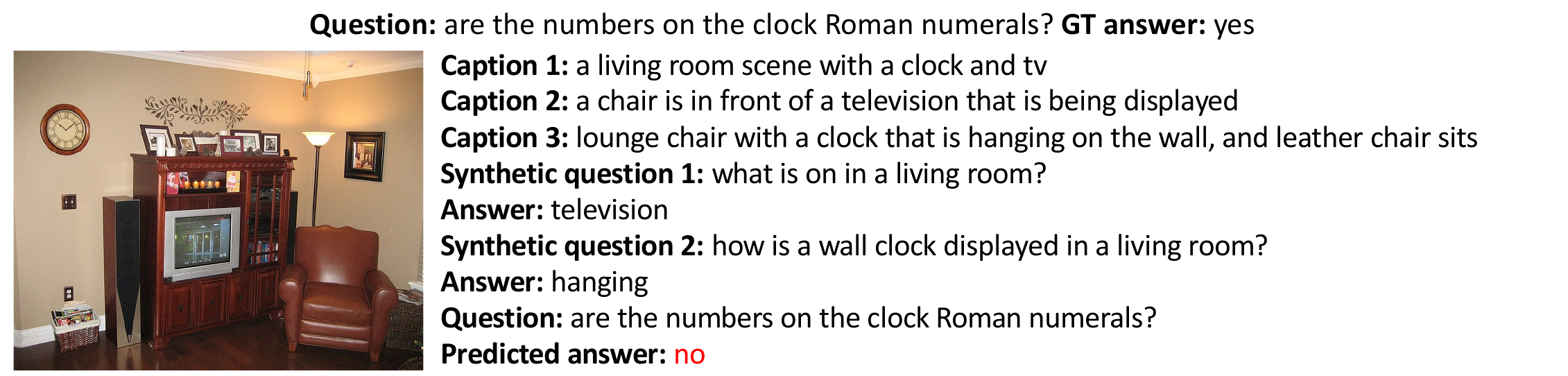}
		\end{minipage}
	}
	\vspace{-0.8em}
	
	\caption{{Failure case analysis for VQAv2. Red color indicates incorrect prediction.} }
	\vspace{-1.em}
\end{figure*}